\documentclass{article}

\PassOptionsToPackage{numbers, compress}{natbib}

\usepackage[preprint]{neurips_2023}

\usepackage[utf8]{inputenc} 
\usepackage[T1]{fontenc}    
\usepackage{url}            
\usepackage{booktabs}       
\usepackage{amsfonts}       
\usepackage{nicefrac}       
\usepackage{microtype}      
\usepackage{xcolor}         
\usepackage{color, soul}
\usepackage{graphicx,wrapfig,lipsum,booktabs}
\usepackage{amsmath,bm}
\usepackage{caption}
\usepackage{subcaption}
\usepackage[ruled,vlined]{algorithm2e}
\usepackage{dsfont}

\usepackage{array,multirow}
\usepackage{amssymb}
\usepackage{tabularx}
\usepackage{hyperref}
\usepackage{algcompatible}
\hypersetup{pdfborder=0 0 0}

\newcommand{\STAB}[1]{\begin{tabular}{@{}c@{}}#1\end{tabular}}

\SetKwComment{COMMENT}{}{}

\title{NESS: Node Embeddings from Static SubGraphs}

\author{Talip Uçar\\
Centre for AI, BioPharmaceuticals R\&D, AstraZeneca\\
\texttt{talip.ucar@astrazeneca.com}
}

\begin{document}

\maketitle

\begin{abstract}
We present a framework for learning \textbf{N}ode \textbf{E}mbeddings from \textbf{S}tatic \textbf{S}ubgraphs (NESS) using a graph autoencoder (GAE) in a transductive setting. NESS is based on two key ideas: i) Partitioning the training graph to multiple static, sparse subgraphs with non-overlapping edges using random edge split during data pre-processing, ii) Aggregating the node representations learned from each subgraph to obtain a joint representation of the graph \textit{at test time}. Moreover, we propose an \textit{optional} contrastive learning approach in transductive setting. We demonstrate that NESS gives a better node representation for link prediction tasks compared to current autoencoding methods that use either the whole graph or stochastic subgraphs. Our experiments also show that NESS improves the performance of a wide range of graph encoders and achieves state-of-the-art results for link prediction on multiple real-world datasets with edge homophily ratio ranging from strong heterophily to strong homophily.
\end{abstract}

\section{Introduction}

Link prediction in graphs is a firmly established problem in the literature, especially in the area of network analysis, and has a wide range of applications in many domains such as social, biological and transportation networks \citep{zhang2020autosf, qi2006evaluation, chami2019hyperbolic}, recommender systems \citep{zhang2018link} and cybersecurity \citep{liben2003link}. The problem is usually tackled by first learning node embeddings that capture both topology and node features of the graph to represent graph data in a low-dimensional latent space. Various methods have been proposed to extract node embeddings such as matrix factorization \citep{ahmed2013distributed, cao2015grarep, katz1953new}, random walk based methods \citep{perozzi2014deepwalk, grover2016node2vec} and graph convolutional networks \citep{kipf2016semi, kipf2016variational,  hamilton2017inductive, chiang2019cluster}. Among them, one of the most popular set of methods to learn node embeddings are graph autoencoders (GAEs) \citep{kipf2016variational}, which are an extension of autoencoders (AEs) \citep{rumelhart1985learning} to the graph domain. Similar to AE, they consist of: i) An encoder, which is usually based on a graph neural network (GNN) to encode local graph structure and features of nodes in the graph, $G$, into a latent representation $\pmb{Z} \in \mathcal{R}^{N \times d}$, where $N$ is the total number of nodes in the graph and $d$ is the feature dimension. ii) A decoder, which is used to reconstruct the original graph from the latent representation, $\pmb{Z} $. Once the model is trained, we can use $\pmb{Z} $ for downstream tasks such as link prediction \citep{kipf2016variational, berg2017graph}, node and graph classification \citep{rong2019dropedge, kipf2016semi}, graph generation \citep{jin2018junction, simonovsky2018graphvae, samanta2020nevae}, and node clustering \citep{hasanzadeh2019semi, pan2018adversarially}.

\begin{figure*}[ht]
\vskip 0.2in
\begin{center}
\centerline{\includegraphics[width=1\textwidth]{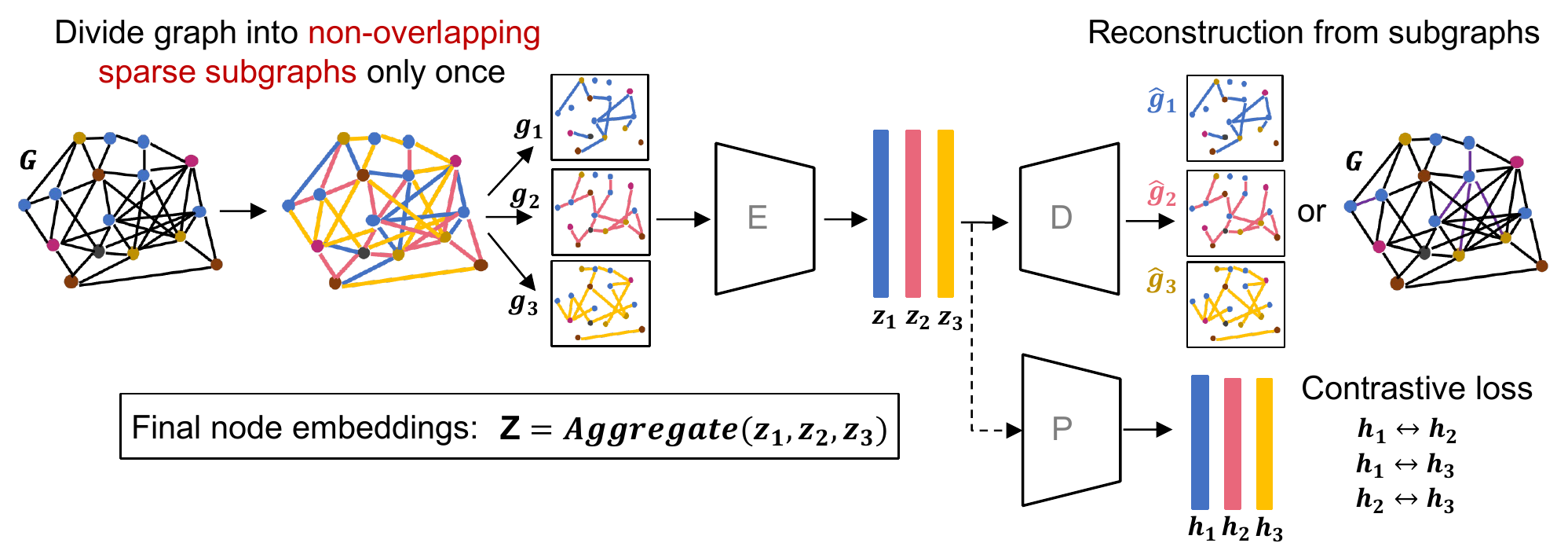}}
\caption{\textbf{NESS framework:} Node embeddings from static subgraphs}\label{fig:subgraph_framework}
\end{center}
\vskip -0.2in
\end{figure*}

Although GNN-based GAEs \citep{kipf2016variational} are effective for learning low dimensional node embeddings, they have a few shortcomings. To begin with, they suffer from scalability issues due to the following reasons \citep{chiang2019cluster}: i) The standard GAE framework proposed in \citep{kipf2016variational} uses a transductive setting and trains the model using full-batch gradient descent, leading to a large memory requirement. ii) The time complexity and computational overhead grow with the number of layers in the encoder due to neighborhood expansion problem \citep{chiang2019cluster}. To compute the loss on a single node at layer $L$, we need the embeddings of its neighbors at the layer $L-1$ and the neighborhood grows as the number of layers increases. iii) Reconstructing a large graph by using the inner-product of node embeddings is computationally costly, O($N^2$) \citep{salha2021fastgae}. Moreover, standard GNN-based GAEs are not expressive enough to extract useful node embeddings for link prediction tasks \citep{pan2021neural, xu2018powerful, morris2019weisfeiler}, for which learning the topology of a graph is important. Although GNN-based encoders capture some knowledge of topology, they are rather limited: when the encoders are shallow (i.e. having few layers), they only capture the immediate neighborhood around the nodes. When they are too deep, new problems such as over-smoothing \cite{li2018deeper, chen2020measuring} and over-squashing \citep{alon2020bottleneck} arise. Over-smoothing is a result of node representations converging to a constant while over-squashing emerges when too many messages are compressed to a fixed length vector \citep{alon2020bottleneck}. Moreover, GAEs are prone to overfitting since their optimisation objective is based on the reconstruction of a large sparse graph, resulting in sub-par performance for link prediction \citep{Goyal_2018}. Finally, most GNN encoders such as GCN \citep{kipf2016semi} are designed under the assumption of homophily and mix the embeddings of neighbouring nodes via averaging their embeddings. Hence, they may not perform well in link prediction task for graphs with strong heterophily \citep{zhu2020beyond}.

Despite its shortcomings, GAEs are a natural way to address link prediction problems since we use a relevant task, i.e. reconstruction of known links, to learn the node embeddings. In recent years, a number of proposals have focused on leveraging subgraphs and various sampling approaches to address the scalability issue in GAEs \citep{hamilton2017inductive, chiang2019cluster, zeng2019graphsaint, salha2021fastgae, chen2018fastgcn, chen2017stochastic}. In contrast, we focus on using subgraphs to improve the predictive performance of node embeddings for link prediction under the transductive setting. Our approach is novel in the following way: i) We first partition the training graph into multiple \textit{static}, sparse subgraphs with non-overlapping edges during data preprocessing step rather than sampling random subgraphs during training. We then train a single GAE using the subgraphs in a transductive setting. ii) At test time, we aggregate the embeddings of subgraphs to obtain the joint embedding for the original graph. The combination of steps (i) and (ii) can be considered as turning the task of trasductive learning on a large graph into the task of multi-view learning \citep{bachman2019learning}, in which each subgraph can be considered a different view of the original graph. One of the benefits of NESS is that working with predefined subgraphs allows us to match the setup during training to the one at test time, especially if the size of the subgraphs is similar to that at test time. The main contributions of this paper can be summarized as:

\begin{itemize}
    \item We propose a simple yet effective framework, NESS, for learning node embeddings from \textit{static subgraphs} in GAE setting and empirically demonstrate its effectiveness on multiple benchmark datasets ranging from low to high edge homophily ratios.
    \item We show that NESS improves the performance of existing GNN-based GAEs for link prediction tasks, achieving SOTA performance on multiple datasets.
    \item We discuss the two principles underlying NESS; i) dividing graphs into non-overlapping, sparse subgraphs, and ii) aggregating embeddings at test time. We give further insights on the issue with mixing node embeddings and compare NESS with ensembling methods. 
    \item We introduce an \textit{optional} contrastive learning approach in transductive learning setting.
    \item We conduct extensive experiments to compare various sampling approaches under GAE setting, giving insights into how they perform across many datasets and encoder types.
\end{itemize}

\section{Method}

\textbf{Definitions} Without the loss of generality, we assume a transductive learning setting throughout this work. We define an undirected, unweighted large graph as $G = (\mathcal{V}, \mathcal{E})$ with vertices $\mathcal{V}$ and edges $\mathcal{E}$. We refer to its adjacency matrix as $\pmb{A} \in \mathcal{R}^{N \times N}$ and collection of its nodes as $\pmb{X} \in \mathcal{R}^{N \times f}$. We assume that diagonal elements of $\pmb{A}$ are set to 1, i.e. including self-loops. For the autoencoder setting, we define the latent vector as $\pmb{Z} \in \mathcal{R}^{N \times d}$. $N=|\mathcal{V}|$ is the total number of nodes in the graph while $f$ and $d$ are the feature dimension of nodes in $\pmb{X}$ and $\pmb{Z}$ respectively. In this work, we use lowercase letters to refer to subgraphs, while , depending on the context, the capital letters are associated with either the original large training graph, or the entire graph that include training, validation and test. Thus, we define $k$ subgraphs, their node representations, adjacency matrices, edges and latent representations as $\{\pmb{g}_1, \pmb{g}_2, ...,\pmb{g}_k\}$, $\{\pmb{x}_1, \pmb{x}_2, ...,\pmb{x}_k\}$, $\{\pmb{a}_1, \pmb{a}_2, ...,\pmb{a}_k\}$, $\{\pmb{e}_1, \pmb{e}_2, ...,\pmb{e}_k\}$ and  $\{\pmb{z}_1, \pmb{z}_2, ...,\pmb{z}_k\}$ respectively. We formally define $k^{th}$ subgraph as $\pmb{g}_k = (\pmb{v}_k, \pmb{e}_k)$ with vertices $\pmb{v}_k$ and edges $\pmb{e}_k$. The adjacency matrix $\pmb{a}_k$ can be considered as a masked version of $\pmb{A}$ while still preserving the self-loops for all nodes: $\pmb{a}_k = \pmb{M}_k \odot \pmb{A}$.

\subsection{Model} 
Figure~\ref{fig:subgraph_framework} presents our framework, NESS. After splitting the large graph into train, validation and test sets, we further divide the training set into subgraphs during data preparation step. Each subgraph is fed to the same encoder (i.e. parameter sharing) to get their corresponding latent representation. Then, we reconstruct adjacency matrix of the same subgraph by using the inner product decoder. Thus, for $k^{th}$ subgraph, we have: \textbf{Encoder:} $\pmb{z}_{k}=E(\pmb{X}, \pmb{a}_k)$, and \textbf{Decoder:} $\Hat{\pmb{a}}_{k}=\sigma(\pmb{z}_{k}\pmb{z}_{k}^T)$. Moreover, since we incorporate an \textit{optional} constrastive loss during training, we have an additional network (referred as P in Figure~\ref{fig:subgraph_framework}) to project the latent variables. It is important to note that since we operate in transductive setting, each $\pmb{z}_k$ corresponding to subgraph $\pmb{g}_k$ actually contains the embeddings of all $N$ nodes of graph $G$ (i.e. $\pmb{z}_k \in \mathcal{R}^{N \times d}$). The embeddings of connected nodes in subgraph $\pmb{g}_k$ are computed by using their neighborhood information while the embeddings for the remaining nodes are computed using only self-loops. When reconstructing the adjacency matrix $\Hat{\pmb{a}}_{k}$, we only need the connected nodes in the subgraph $\pmb{g}_k$ to compute scores for links in $\pmb{g}_k$. From Figure~\ref{fig:subgraph_framework}, we see that training the model with a static subgraph emulates the same transductive learning setting as the one proposed in \citep{kipf2016variational} using the larger graph. Hence, NESS turns the standard transductive setting of GAE to multiple transductive settings, one for each subgraph.

\subsection{Training and Test time}
In this work, we focus on autoencoder setting, thus we use reconstruction loss as our main training objective. However, we also experiment with contrastive learning in NESS framework to improve the performance of our proposed method for various encoder types such as GCN \citep{kipf2016semi}, GAT \citep{velivckovic2017graph}, and ARGA \citep{pan2018adversarially} although it is optional. Our training objective function is:
\begin{equation} \label{eq2}
\mathcal{L}_t =\mathcal{L}_r+\alpha \mathcal{L}_c 
\end{equation}
where $\mathcal{L}_t$, $\mathcal{L}_r$, and $\mathcal{L}_c$ are the total, reconstruction and contrastive losses respectively, and $\alpha \in \{0,1\}$. We don't use contrastive learning (i.e. $\alpha=0$), unless explicitly stated.

\textbf{Reconstruction loss} Given a subgraph $\pmb{g}_k$, we can reconstruct either the same subgraph, $\pmb{\Hat{g}}_k$, or the larger training graph $\pmb{\Hat{G}}$. 
Then, we can compute the reconstruction loss for $k^{th}$ subgraph by computing binary cross-entropy loss using either the subgraph and its reconstructed counterpart $(\pmb{a}_k, \pmb{\Hat{a}}_k)$, or the large graph and the corresponding reconstructed graph $(\pmb{A},\pmb{\Hat{A}}_k)$ pair as shown in Figure \ref{fig:subgraph_framework}. We choose the former, i.e. $(\pmb{a}_k, \pmb{\Hat{a}}_k)$, since it is either on par or more effective in our experiments as shown in Figure~\ref{fig:ness_ablation1}b. Reconstruction loss is computed for both positive and negative edges, where we obtain negative edges by randomly sampling same number of negative edges as positive ones: 
\begin{equation}\label{eq3}
\begin{aligned}
&\mathcal{L}_r = \frac{1}{K}\sum_{k=1}^{K} l_k \mbox{, \, where }
l_k = -\frac{1}{2n_{k_p}} &\sum_{i=1}^{n_{k_p}} [ \log \pmb{\Hat{e}}_{ki} + \log(1-\lnot\pmb{\Hat{e}}_{ki})]\,,
 \end{aligned}
\end{equation}
where $\mathcal{L}_r$ is the average of reconstruction losses $l_k$ over all $k=1,\ldots,K$ subgraphs, $n_{k_p}$ is the number of positive edges in the edge set $\pmb{e}_k$ while $\pmb{\Hat{e}}_{ki}$ and $\lnot\pmb{\Hat{e}}_{ki}$ refer to the scores for $i^{th}$ positive and sampled negative edge in $k^{th}$ subgraph respectively.


\textbf{Contrastive loss}
Referring to the example in Figure~\ref{fig:subgraph_framework}, since we are in transductive setting, we have three different views for each node in the graph: \{$\bm{z}_1$, $\bm{z}_2$, $\bm{z}_3$\}. Nodes in the same index position across three latent vectors become positive samples of the node while the nodes in other indexes can be considered as negative samples for that particular node. Therefore, we can compute a constrastive loss between any two pairs from the set of all latent vectors \{$\bm{z}_1$, $\bm{z}_2$, $\bm{z}_3$\}. Moreover, instead of using \{$\bm{z}_1$, $\bm{z}_2$, $\bm{z}_3$\} directly, we first project them using a projection network (P) to obtain \{$\bm{h}_1$, $\bm{h}_2$, $\bm{h}_3$\} \citep{chen2020simple}. Thus, we compute the normalized temperature-scaled cross entropy loss (NT-Xent) \citep{chen2020simple} for every pair $\{\bm{h}_a, \bm{h}_b\}$ of total three pairs from the set $S=\{\{\bm{h}_1, \bm{h}_2\},\{\bm{h}_1, \bm{h}_3\},\{\bm{h}_2, \bm{h}_3\}\}$. Overall contrastive loss is: 
\begin{equation}\label{eq4}
\mathcal{L}_c = \frac{1}{J}\sum_{\{\bm{h}_a,\bm{h}_b\}\in S}p(\bm{h}_a,\bm{h}_b) \mbox{, where  } \resizebox{0.55\textwidth}{!}{$p(\bm{h}_a,\bm{h}_b)=\frac{1}{2N}\sum_{i=1}^{N}\left[l({\bm{h}_a}^{(i)}, {\bm{h}_b}^{(i)}) + l({\bm{h}_b}^{(i)}, {\bm{h}_a}^{(i)})\right]$}
\end{equation} 
\begin{equation}\label{eq6}
\resizebox{0.55\textwidth}{!}{$l({\bm{h}_a}^{(i)}, {\bm{h}_b}^{(i)})=-\log\frac{\exp(sim({\bm{h}_a}^{(i)},{\bm{h}_b}^{(i)})/\tau)}{\sum_{k=1}^{2N} \mathds{1}_{k \neq i} \exp(sim({\bm{h}_a}^{(i)}, \bm{{h_b}^{(k)}})/\tau)}$}\,,
\end{equation} 
where $J$ is the total number of pairs in set $S$, $\tau$ is the temperature coefficient, $p(\bm{h}_a,\bm{h}_b)$ is total contrastive loss for a pair of projection $\{\bm{h}_a, \bm{h}_b\}$, $l({\bm{h}_a}^{(i)}, {\bm{h}_b}^{(i)})$ is the loss function for a corresponding positive pairs of nodes \{${\bm{h}_a}^{(i)}, {\bm{h}_b}^{(i)}$\} in subgraphs $\{\bm{h}_a, \bm{h}_b\}$, and $\mathcal{L}_c$ is the average of contrastive losses over all pairs. 

\textbf{Test time}
At test time, we use an aggregation function to obtain the joint embedding of the graph $G$ from the subgraphs: $\bm{Z}=\frac{1}{K}\sum_{k=1}^{K} \bm{z}_k$. We can use any permutation invariant aggregation method, including mean, sum, min, and max. We use mean aggregation in all our experiments. The NESS framework is summarized as pseudocode in Algorithm~\ref{alg:ness} in the Appendix.

\subsection{The principles behind NESS:}\label{ness_principles}
The NESS is based on two main principles: dividing the graph into sparse subgraphs and aggregating embeddings at test time. The former reduces the direct mixing of node representations, which benefits heterophilous graphs while the latter allows learning the global structure through multiple subgraphs, improving performance for homophilous graphs \citep{zhu2020beyond}.

NESS treats each node's local neighborhood in different subgraphs as multiple views for link prediction, enabling the model to leverage information from different perspectives. This can also be considered as a form of ensemble learning, which diversifies predictions and improves accuracy. For example, assuming that we have two subgraphs and that we use DistMult \citep{yang2014embedding} as our scoring function for link prediction between two nodes (e.g., node 1 and 2), we can take the linear approximation of the sigmoid function, $\sigma(x)$, around the decision threshold $x\approx0$ and express the likelihood of a link existing between two nodes as:
\begin{equation}
\hat{y}=\sigma({\bm{z}_1}^T{\bm{z}_2}) \approx \frac{1}{2} + \frac{{\bm{z}_1}^T{\bm{z}_2}}{4} = \frac{1}{2} + \frac{{\bm{z}_{1_1}}^T\bm{z}_{2_1}+{\bm{z}_{1_1}}^T\bm{z}_{2_2}+{\bm{z}_{1_2}}^T\bm{z}_{2_1}+{\bm{z}_{1_2}}^T\bm{z}_{2_2}}{16}\,,
\end{equation}
where we express $\bm{z}_1$ as the average of two embeddings, \{$\bm{z}_{1_1}$, $\bm{z}_{1_2}$\} for node 1 in two subgraphs, and we do the same for node 2. So the final score is computed as the average of four dot product terms between two nodes, with two terms from node embeddings extracted from the same subgraphs and the other two from different subgraphs. The inter- and intra-subgraph dot products help diversify predictions, contributing to better accuracy. However, to achieve better accuracy than individual subgraphs, there should be sufficient dissimilarity between subgraphs \citep{ho1998random}. The difference in accuracy obtained from aggregated embeddings and individual subgraphs can give indirect evidence of mutual independence between the subgraphs, which we explore in Section~\ref{results}.

\section{Experiments}\label{exp_setup}

\textbf{Datasets} We use three standard benchmark datasets from citation networks (Cora, Citeseer and Pubmed) \citep{sen2008collective}, where nodes and edges correspond to documents and undirected citations respectively. Node features are represented as bag-of-words representations of a document. These datasets are considered the homophilous graphs, where nodes tend to connect with similar other nodes. \citep{mcpherson2001birds}. We also use three datasets (Cornell, Texas, Wisconsin) from WebKB \citep{pei2020geom}, which include web pages from computer science departments of multiple universities. Here node features are the bag-of-words representation of web pages. Finally, we consider the Chameleon dataset \citep{rozemberczki2021multi}, which is a wikipedia page-page graph under the topic chameleon. The graphs from WebKB and Chameleon are considered to be heterophilous, hence connected nodes are prone to having different properties or labels \citep{zhu2020beyond, wang2022augmentation}. Statistics of all datasets are summarised in Table~\ref{data_stats} in the Appendix.

\textbf{Data pre-processing} Following \citep{kipf2016variational}, we split the graph into three parts: 10\% testing, 5\% validation, 85\% training, using random edge-level splits (RES). We then perform RES to generate $k$ subgraphs from training set, such that subgraphs do not share any edges. We use $k \in [2, 4, 8]$ in our experiments.

\textbf{Models} Since NESS is general to any GNN-based autoencoder, we can use any GNN-based encoder as our backbone to obtain node representations. Different encoders adopt different mechanisms to learn from both graph topology and node features, resulting in different node representations. In our experiments, we use GCN \citep{kipf2016variational}, GAT \citep{velivckovic2017graph}, GNAE \citep{ahn2021variational}, as well as models such as ARGA \citep{pan2019learning} that use adversarial methods. We also use a linear version of GCN-based encoder, referred as Lin. All encoders are shallow (one or two layers) with the final layer dimension of 32. We give further details of models in Appendix~\ref{details_of_models}.


\textbf{Baselines} The NESS method is designed for GAEs in transductive learning setting, so we use two other frameworks as our main baselines: i) The standard GAE framework from \citep{kipf2016variational}, where the input and output of the model is the entire training graph. We refer to this setup as SGAE when reporting results. ii) The FastGAE framework from \citep{salha2021fastgae}, where the input to the model is the entire training graph, and the model reconstructs dynamically sampled subgraphs at the output. We use our own random edge splitting (RES) method to generate subgraphs instead of the node sampling method used by FastGAE, to make it easier to compare different GAE setups. For FastGAE, we sample 50\% of the training graph per epoch when generating subgraphs during training. We also report the original results for the link prediction task from FastGAE.

As a variation of the original FastGAE setting, we experiment with a setup where the encoder encodes a dynamically sampled subgraph, while the decoder tries to reconstruct the same subgraph. We use the RES method to sample a subgraph per each iteration, with three dynamic sampling rates: 50\%, 75\%, and 88\% of the training graph. We refer to them as DS50, DS75, and DS88 respectively.

To test whether using RES partitioning during data preprocessing matters (i.e. generating static subgraphs), we also experiment with its stochastic counterpart: We partition the training graph into $k$ subgraphs dynamically by using RES during training. The main difference is that we sample $k$ new subgraphs every iteration (i.e. not static). We generate two random subgraphs per iteration, each of which is 50\% of the training graph. We refer to this setting as dynamic random edge splitting with 2 subgraphs (DRES2). Note that DRES2 partitions graph into two subgraphs per iteration without missing any training edges while DS50 generates one subgraph using 50\% of edges per iteration.

For NESS, we experiment with 2, 4, and 8 static subgraphs and refer them as NESS2, NESS4, and NESS8 in our experiments respectively. We also randomly drop edges ($p=0.2$) for both NESS and SGAE to reduce overfitting as the graphs are static in both cases. For all baselines and NESS, we use the same setup and pipeline, with the only differences being whether we use the whole graph, stochastic subgraphs, or static subgraphs. We compare all of these using multiple encoders and benchmark datasets. The summary of all settings is in Table~\ref{summary_settings}.

\begin{wraptable}{R}{0.5\textwidth}
\caption{Summary of settings for the baselines and NESS. ($\pmb{\Bar{a}}_k, \pmb{\Bar{z}}_k$) in NESS are deterministic while ($\pmb{\Tilde{a}}_k, \pmb{\Tilde{z}}_k$) indicate that we use sampling during training. $k$ indicates the $k^{th}$ subgraph in settings where multiple subgraphs per epoch are used in training.}
\label{summary_settings}
\vskip 0.15in
\begin{center}
\begin{small}
\begin{sc}
\resizebox{0.48\columnwidth}{!}{
{\begin{tabular}{lcccr}
\toprule
Method & Encoding & Decoding & $\pmb{Z}$ at test time \\
\midrule
SGAE    & $\pmb{Z}=E(\pmb{X}, \pmb{A})$ & $\pmb{\Hat{A}}=\sigma(\pmb{Z}\pmb{Z}^T)$&   $\pmb{Z}=E(\pmb{X}, \pmb{A})$ \\
FGAE    & $\pmb{Z}=E(\pmb{X}, \pmb{A})$ & $\pmb{\Hat{a}}=\sigma(\pmb{\Tilde{z}}\pmb{\Tilde{z}}^T)$&   $\pmb{Z}=E(\pmb{X}, \pmb{A})$  \\
DS50    & $\pmb{\Tilde{z}}=E(\pmb{X}, \pmb{\Tilde{a}})$ & $\pmb{\Hat{a}}=\sigma(\pmb{\Tilde{z}}\pmb{\Tilde{z}}^T)$& $\pmb{Z}=E(\pmb{X}, \pmb{A})$  \\
DRES2     &${\pmb{\Tilde{z}}_k}=E(\pmb{X}, {\pmb{\Tilde{a}}_k})$ & $\pmb{\Hat{a}_k}=\sigma({\pmb{\Tilde{z}}_k}{\pmb{\Tilde{z}}_k}^T)$ & $\pmb{Z}=E(\pmb{X}, \pmb{A})$ \\
NESS      & ${\pmb{\Bar{z}}_k}=E(\pmb{X}, {\pmb{\Bar{a}}_k})$ & $\pmb{\Hat{a}_k}=\sigma({\pmb{\Bar{z}}_k}{\pmb{\Bar{z}}_k}^T)$&   $\pmb{Z}=Agg(\pmb{\Bar{z}_1}, ...,{\pmb{\Bar{z}}_k})$      \\
\bottomrule
\end{tabular}}{}
}
\end{sc}
\end{small}
\end{center}
\vskip -0.1in
\end{wraptable}

\textbf{Training and optimisation} We use AdamW optimizer \citep{loshchilov2017decoupled} with a learning rate of 0.01, $betas=(0.9, 0.999)$ and $eps=1e-07$ for all of our experiments. We set the maximum number of epochs as 500, and use early stopping by using validation set loss with patience of $[3,15]$ epochs range. We pick the model with the smallest validation loss.

\textbf{Evaluation} Following \citep{kipf2016variational}, we evaluate link prediction by measuring the area under the ROC curve (AUC) and average precision (AP) scores on the test set. For all experiments, we repeat each experiment ten times with different random seeds (i.e. train-val-test splits and model initialisation). The code for NESS is provided at \url{https://github.com/AstraZeneca/NESS}

\begin{figure*}[t]
\vskip 0.2in
\begin{center}
     \begin{subfigure}[c]{0.19\textwidth}
         \includegraphics[width=\textwidth]{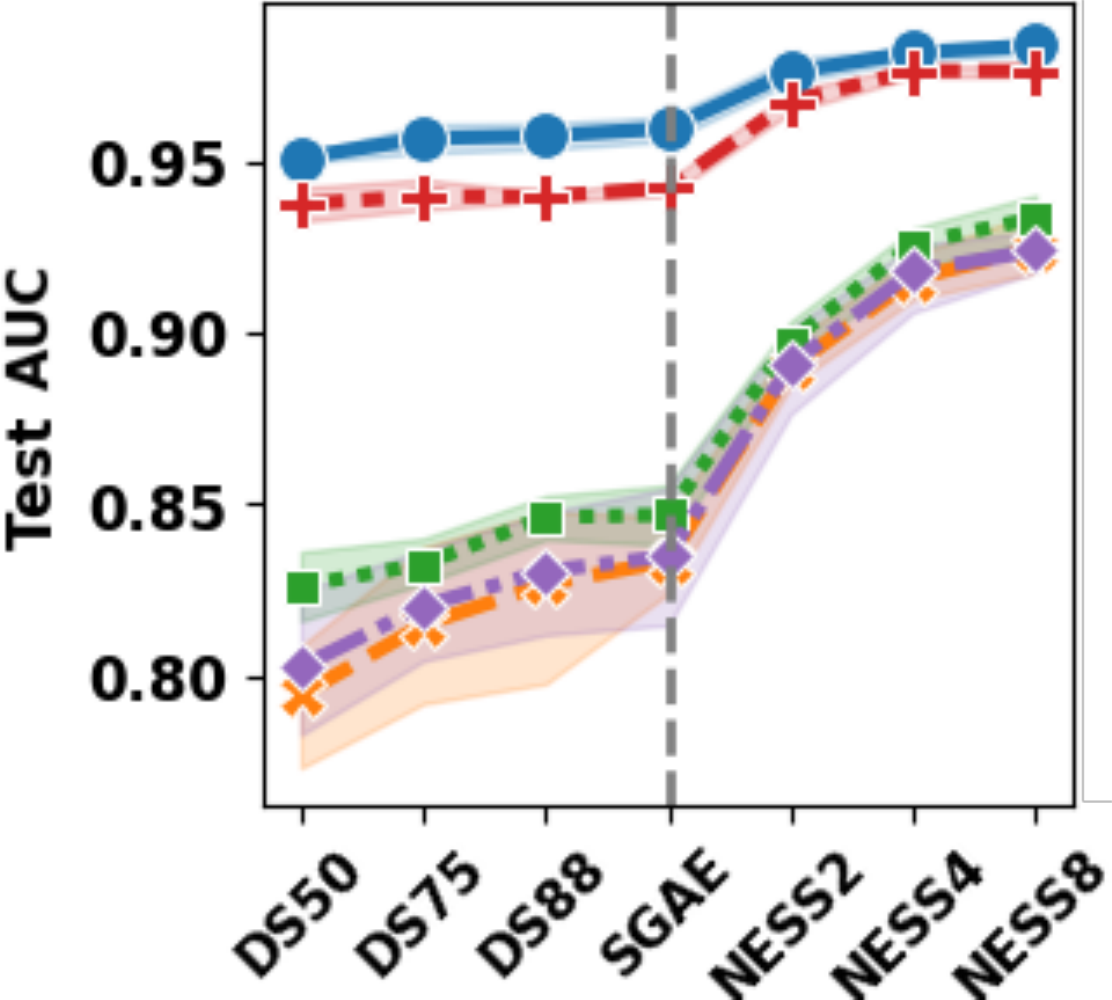}
         \caption{Cora}
     \end{subfigure}
     \begin{subfigure}[c]{0.19\textwidth}
         \includegraphics[width=\textwidth]{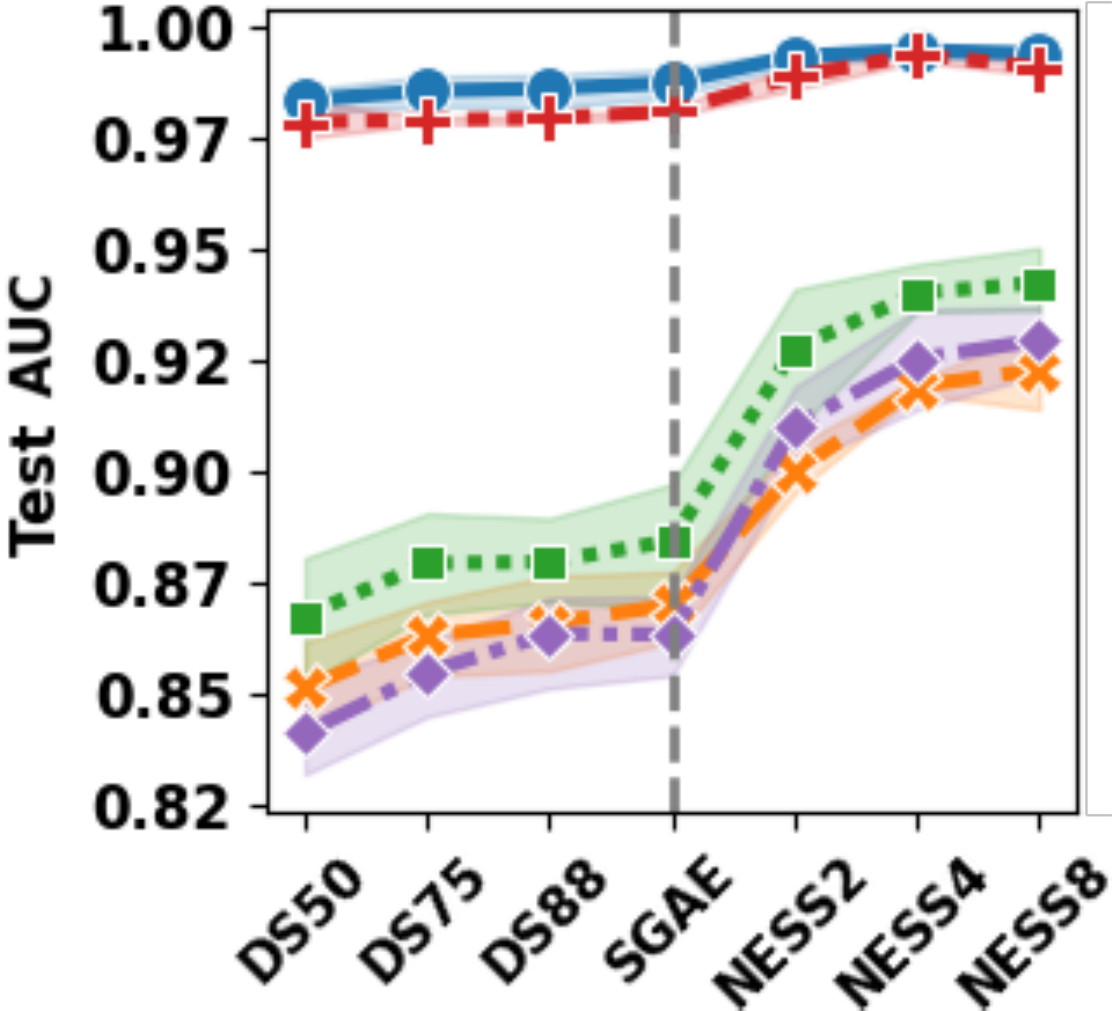}
         \caption{Citeseer}
     \end{subfigure}
     \begin{subfigure}[c]{0.19\textwidth}
         \includegraphics[width=\textwidth]{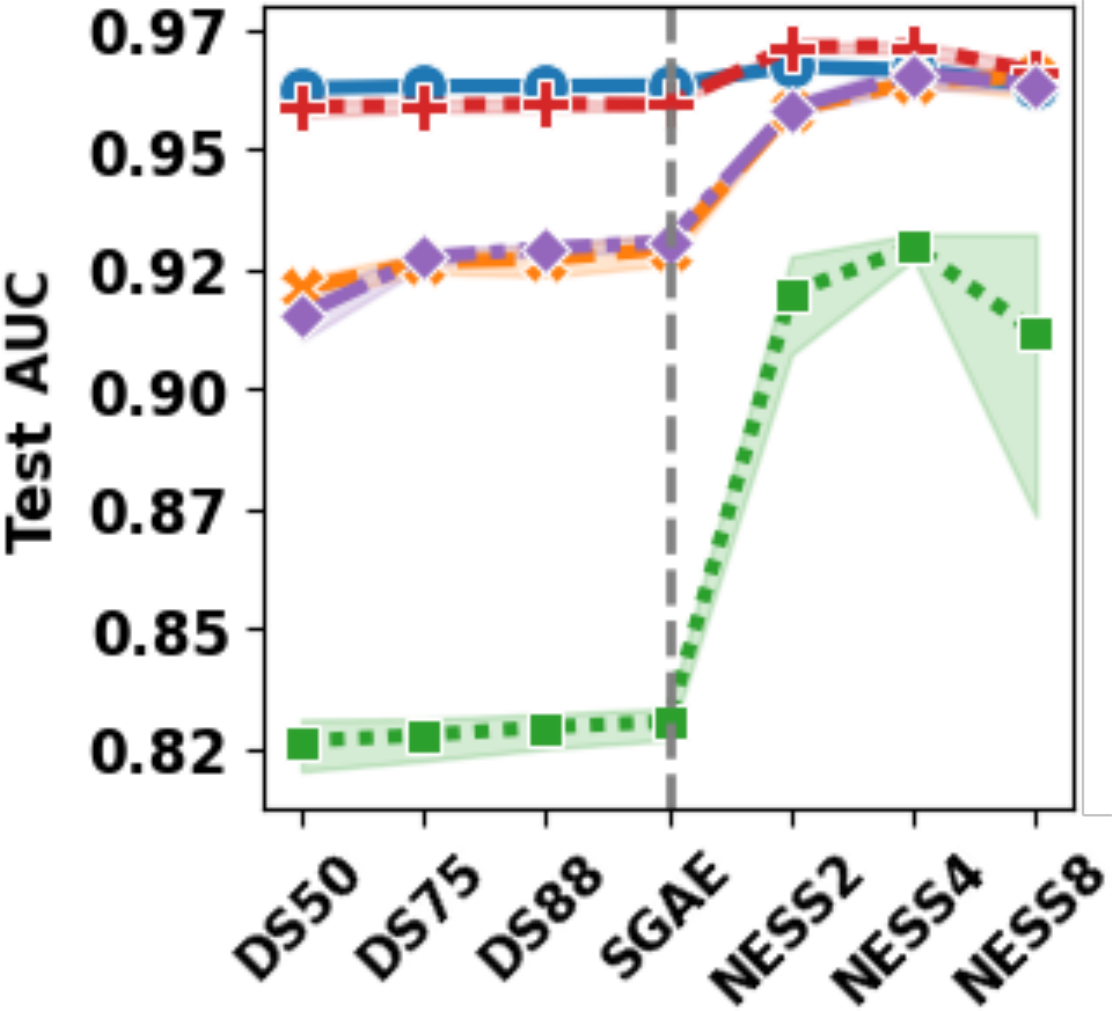}
         \caption{Pubmed}
     \end{subfigure}
     \begin{subfigure}[c]{0.08\textwidth}
         \includegraphics[width=\textwidth]{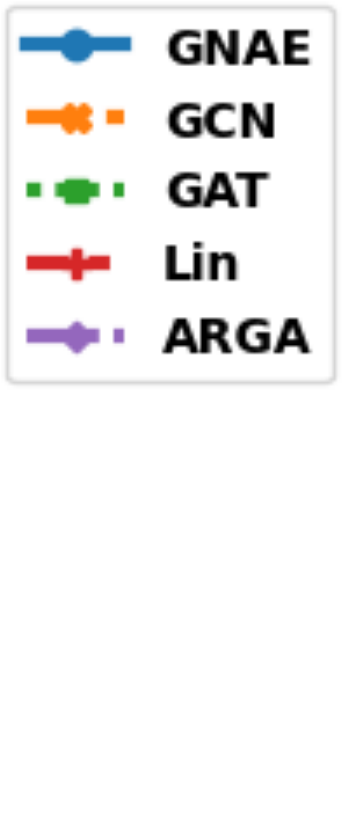}
     \end{subfigure}
     \begin{subfigure}[c]{0.19\textwidth}
         \includegraphics[width=\textwidth]{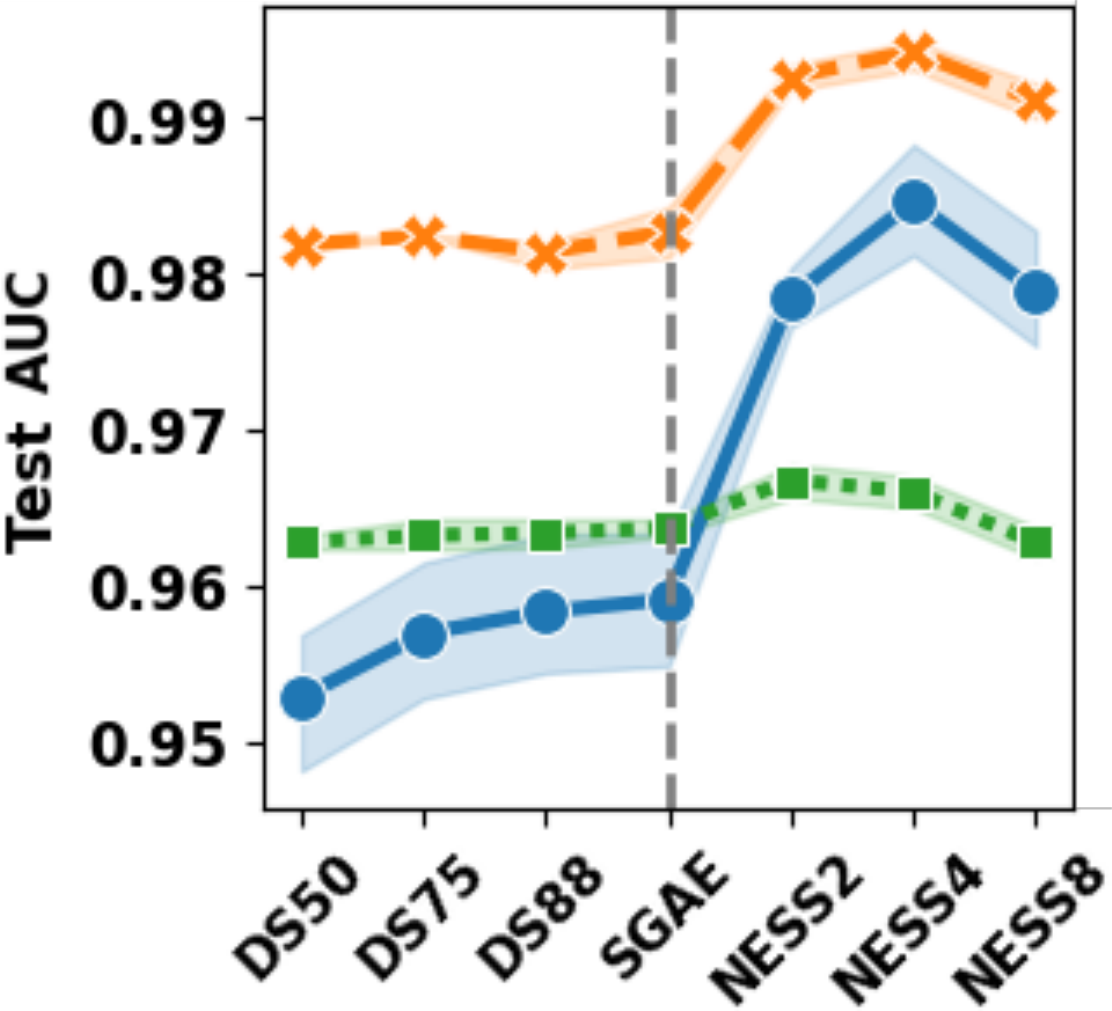}
         \caption{GNAE}
     \end{subfigure}
     \begin{subfigure}[c]{0.12\textwidth}
         \includegraphics[width=\textwidth]{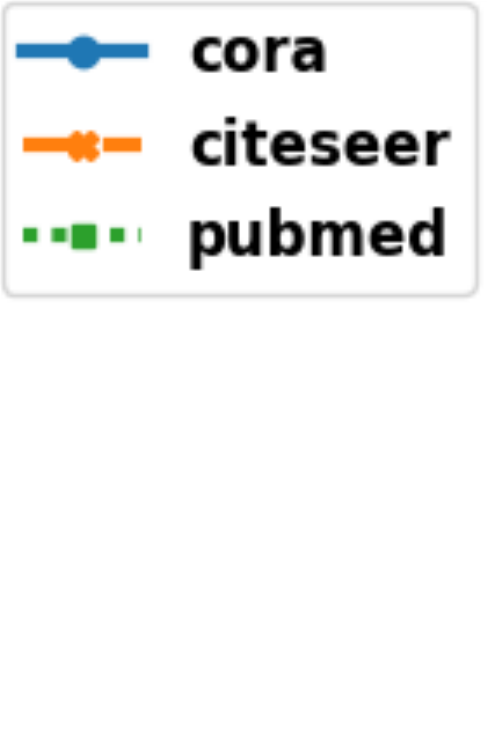}
     \end{subfigure}     
\caption{Comparing NESS to standard GAE (SGAE) and some of the other settings described in Section~\ref{exp_setup}.  Compared to baseline methods, NESS gives a significant performance boost across all encoder types and datasets for link prediction task. \textbf{a-c)} Listing results for multiple encoder types for Cora, Citeseer and Pubmed respectively. \textbf{d)} The performance for GNAE encoder across all settings for all three datasets in detail.}\label{fig:comparing_dynamic_to_static}
\end{center}
\vskip -0.2in
\end{figure*}

\section{Results}\label{results}
\textbf{Comparing NESS to other settings:} We train GAEs with five encoder types (GCN, GAT, GNAE, Lin, ARGA) on seven datasets across the following settings: i) NESS, ii) Standard GAE (SGAE) and iii) Dynamically sampled subgraphs (DS50, DS75, DS88). When training NESS, we use \textit{only the reconstruction loss} for a fair comparison. Figure~\ref{fig:comparing_dynamic_to_static}(a-c) shows performance for citation networks. Key observations include: \textbf{i)} Dynamically sampling larger subgraphs improves performance, approaching SGAE training. \textbf{ii)} NESS outperforms other settings, with performance increasing with more subgraphs. This trend holds across different encoders. We also compare NESS with FastGAE and DRES2 in Table~\ref{table_link_prediction}, which presents a comprehensive comparison of each setting using GNAE as the encoder, except for FGAE1 (GCN). WalkPooling (WP) results are also included. To the best of our knowledge, NESS with GNAE achieves new state-of-the-art performance on five datasets: Cora, Citeseer, and all three WebKB datasets.

\textbf{Properties of subgraphs in NESS:}
For the following analysis, we refer the reader to the discussion in Section~\ref{ness_principles} and Figure~\ref{fig:subgraph_framework}. Using the Cora dataset and five distinct GNN encoders, we examine three cases where the graph is divided into $K=2, 4,$ and $8$ subgraphs.

\begin{figure}[t]
\vskip 0.2in
\begin{center}
     \begin{subfigure}[c]{0.32\columnwidth}
        \includegraphics[width=\columnwidth]{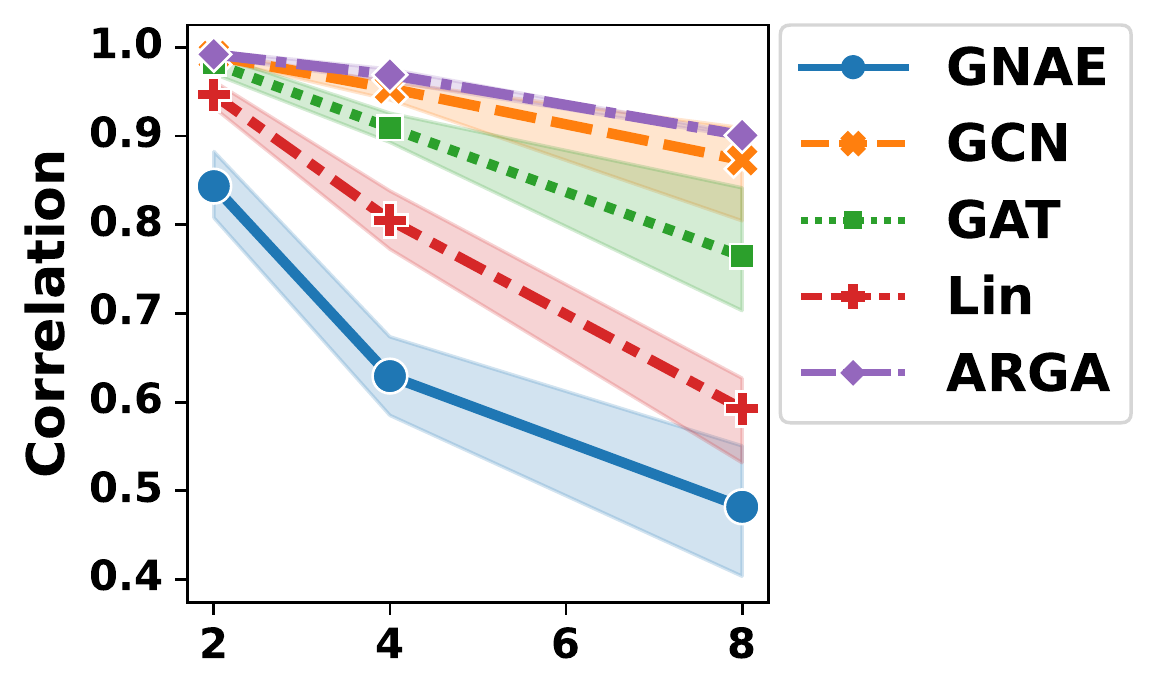}
         \caption{Correlation between subgraphs}
     \end{subfigure}
     \begin{subfigure}[c]{0.32\columnwidth}
        \includegraphics[width=\columnwidth]{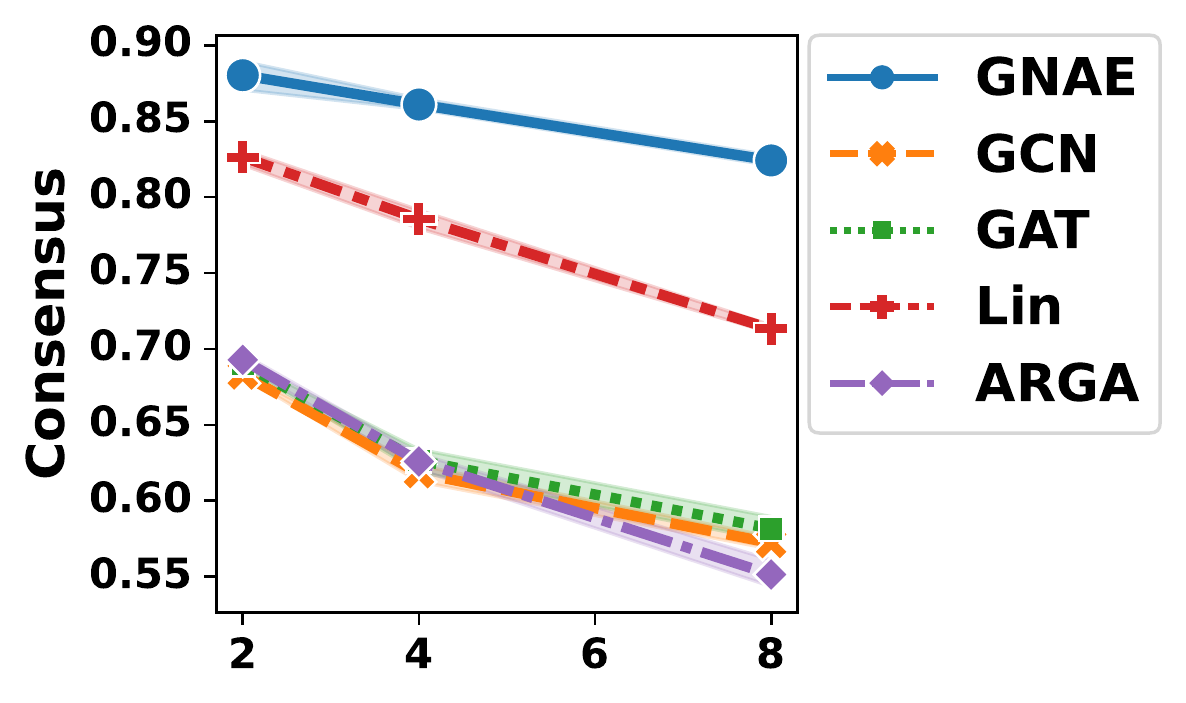}
         \caption{Consensus among subgraphs}
     \end{subfigure}
     \begin{subfigure}[c]{0.32\columnwidth}
        \includegraphics[width=\columnwidth]{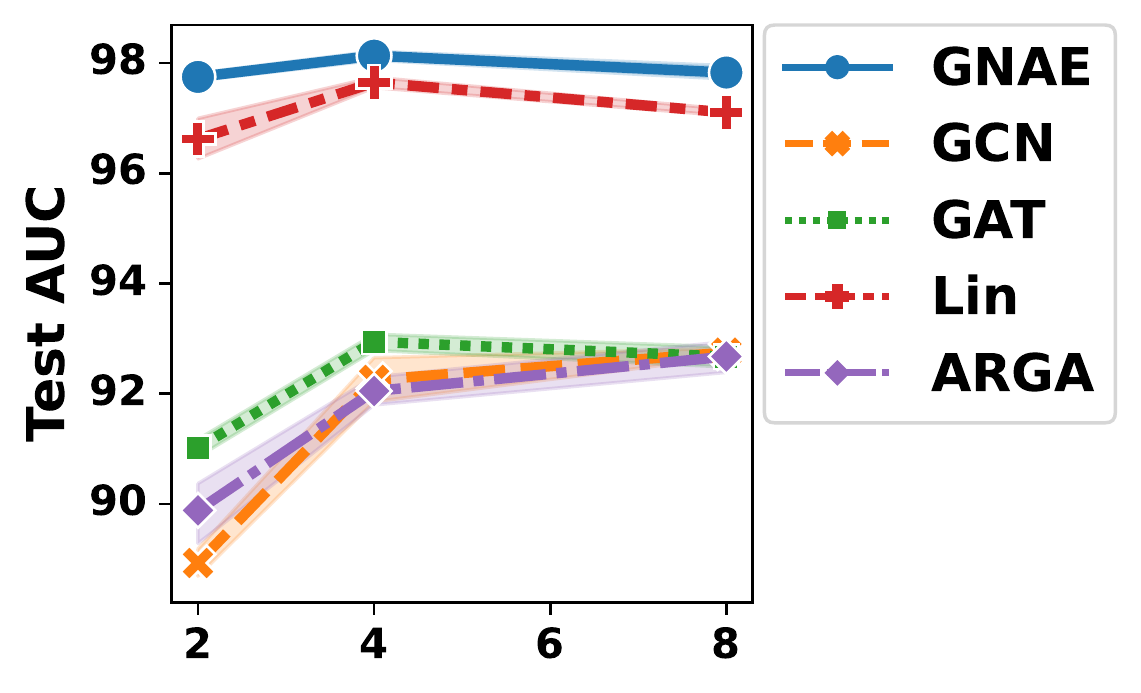}
         \caption{Test AUC using joint embedding}
     \end{subfigure}
\caption{\textbf{Analysis of subgraphs:} \textbf{a)} Correlation between the representations of subgraphs,  \textbf{b) Consensus:} The ratio of links predicted correctly by all subgraphs in the test set,  \textbf{c)} Test AUC using the joint embeddings. X-axis corresponds to three cases; $K=2, 4$ and $8$.}\label{fig:analysis_subgraphs}
\end{center}
\vskip -0.2in
\end{figure}

\textbf{i)} In the first experiment, we first compute the joint embedding of subgraphs for each of the three cases ($K=2, 4,$ and $8$). Next, we extract the embeddings of connected nodes in each subgraph and compute the representation for each subgraph by averaging the embedding of all connected nodes (e.g., for the $k^{th}$ subgraph: ${\bm{g}_k}_{ave} = \frac{1}{n_k}\sum_{i=1}^{n_k} \bm{n}_i : \bm{n}_i \in \bm{g}_k,  n_k=|\bm{g}_k| \}$). Finally, we calculate the Pearson correlation between the subgraph representations, as depicted in Figure~\ref{fig:analysis_subgraphs}a. We observe that the correlation between subgraphs decreases as the graph is divided into more subgraphs. This is expected, as dividing the graph into more sparse subgraphs with non-overlapping edges reduces the mixing and correlation between nodes. This desirable property leads to improved performance in the ensemble of similarity scores computed for pairs of nodes, as discussed in Section~\ref{ness_principles}.

\textbf{ii)} To further investigate the mutual independence of subgraphs, we measure the consensus of link predictions obtained using the latent variables of each subgraph. For instance, for $K=2$, we obtain two separate predictions on the test set from ${\bm{z}_1}$ and ${\bm{z}_2}$, each representing a different subgraph. We then compute the consensus by examining the ratio of common samples predicted correctly by both subgraphs. This process is repeated for $K=4$ and $K=8$. As shown in Figure~\ref{fig:analysis_subgraphs}b, the consensus decreases as $K$ increases, indicating that the subgraphs become more uncorrelated.

\textbf{iii)} Finally, we compute the AUC on the test set using the aggregated embeddings for all three cases, as displayed in Figure~\ref{fig:analysis_subgraphs}c. The results demonstrate that as subgraphs become more uncorrelated (see figures in (a) and (b)), the performance obtained from the joint embedding increases.

\begin{table*}[tb]
\caption{Link prediction results across benchmark datasets for NESS and other settings for GNAE as encoder. NESS outperforms other settings in our experiments on all datasets and the most recent SOTA model WalkPooling (WP) on five out of seven datasets, achieving new SOTA peformance \cite{pan2021neural}. FGAE* and WP* refer to the best reported numbers in the original works. The results from our own implementation of FastGAE with GCN encoder (FGAE1) and with GNAE (FGAE2) are also listed for comparison. For NESS, \textit{the only reconstruction loss} is used during training.}
\label{table_link_prediction}
\vskip 0.15in
\begin{center}
\begin{small}
\begin{sc}
\resizebox{1.0\textwidth}{!}{
{\begin{tabular}{lllllllllllllll}
\toprule
               & \multicolumn{2}{c}{\textbf{chameleon}}                             & \multicolumn{2}{c}{\textbf{citeseer}}                              & \multicolumn{2}{c}{\textbf{cora}}                                  & \multicolumn{2}{c}{\textbf{cornell}}                               & \multicolumn{2}{c}{\textbf{pubmed}}                                & \multicolumn{2}{c}{\textbf{texas}}                                 & \multicolumn{2}{c}{\textbf{wisconsin}}                             \\
\midrule
& \multicolumn{1}{c}{\textbf{AP}} & \multicolumn{1}{c}{\textbf{AUC}} & \multicolumn{1}{c}{\textbf{AP}} & \multicolumn{1}{c}{\textbf{AUC}} & \multicolumn{1}{c}{\textbf{AP}} & \multicolumn{1}{c}{\textbf{AUC}} & \multicolumn{1}{c}{\textbf{AP}} & \multicolumn{1}{c}{\textbf{AUC}} & \multicolumn{1}{c}{\textbf{AP}} & \multicolumn{1}{c}{\textbf{AUC}} & \multicolumn{1}{c}{\textbf{AP}} & \multicolumn{1}{c}{\textbf{AUC}} & \multicolumn{1}{c}{\textbf{AP}} & \multicolumn{1}{c}{\textbf{AUC}} \\
\midrule
\textbf{WP*}& \multicolumn{1}{c}{-}  & \textbf{99.52±0.1}   & \multicolumn{1}{c}{-}    & 95.94±0.5  & \multicolumn{1}{c}{-}   & 95.90±0.5 & \multicolumn{1}{c}{-}  & 82.39±8.9  & \multicolumn{1}{c}{-}     & \textbf{98.72±0.1}  & \multicolumn{1}{c}{-}   & 76.02±7.1   & \multicolumn{1}{c}{-}  & 82.27±6.3     
\multirow{3}{*}{\STAB{\rotatebox[origin=c]{90}{}}} \\
\textbf{FGAE*}& \multicolumn{1}{c}{-}  & \multicolumn{1}{c}{-}   & \multicolumn{1}{c}{90.16±1.2}    & 90.22±1.1  & \multicolumn{1}{c}{92.36±1.1}   & 91.72±1. & \multicolumn{1}{c}{-}  & \multicolumn{1}{c}{-}  & \multicolumn{1}{c}{96.35±0.2}     & 96.12±0.2  & \multicolumn{1}{c}{-}   &  \multicolumn{1}{c}{-}   & \multicolumn{1}{c}{-}  &  \multicolumn{1}{c}{-}     \\
\midrule
\textbf{FGAE1}& 92.80±0.7                       & 93.07±0.7                       & 85.82±1.9                      & 84.47±1.6                       & 82.13±1.3                       & 82.28±1.3                        & 68.17±5.0                      & 69.48±4.7                       & 92.83±0.4                       & 92.61±0.6                       & 61.21±4.3                      & 64.75±1.3                       & 73.75±8.0                      & 74.51±6.6                      \\
\textbf{FGAE2}& 96.92±0.3                      & 96.87±0.2                        & 98.94±0.1                      & 98.76±0.2                       & 96.81±0.6                      & 96.3±0.4                        & 82.05±0.6                      & 86.32±1.4                       & 96.33±0.1                      & 96.32±0.1                       & 83.94±1.9                      & 87.0±2.4                        & 93.68±3.9                      & 93.64±3.4                        \\
\textbf{DS50}& 96.62±0.3                      & 96.54±0.2                       & 98.4±0.0                       & 98.19±0.0                       & 95.93±0.4                      & 95.28±0.5                       & 81.05±5.2                      & 82.76±4.3                        & 96.28±0.1                      & 96.28±0.1                       & 82.92±1.5                      & 87.28±1.6                       & 94.90±1.5                       & 94.61±0.9                       \\
\textbf{DS75}& 96.85±0.3                      & 96.81±0.2                       & 98.46±0.1                      & 98.25±0.0                       & 96.20±0.5                       & 95.69±0.5                       & 80.91±1.5                     & 84.88±1.6                       & 96.31±0.1                      & 96.32±0.1                      & 82.59±1.6                      & 86.21±2.6                       & 92.68±2.1                      & 92.93±1.8                       \\
\textbf{DS88}& 96.90±0.3                       & 96.89±0.2                       & 98.39±0.1                      & 98.14±0.1                       & 96.26±0.6                      & 95.83±0.5                       & 81.64±0.6                      & 85.64±1.3                       & 96.33±0.1                      & 96.34±0.1                       & 81.48±2.2                    & 84.81±2.7                       & 92.33±2.3                      & 93.15±1.6   \\                   
\textbf{SGAE}& 96.93±0.2                      & 96.98±0.1                      & 98.55±0.2                      & 98.27±0.2                        & 96.35±0.6                      & 95.91±0.3                       & 81.13±0.8                       & 85.68±1.2                       & 96.36±0.1                     & 96.37±0.1                       & 81.51±2.0                      & 84.16±2.2                        & 91.78±3.6                       & 92.11±3.3                       \\
\midrule
\textbf{DRES2}& 97.20±0.3     & 97.27±0.2                       & 98.53±0.3                      & 98.4±0.3                       & 96.86±0.6                      & 96.60±0.4                       & 82.04±4.9                      & 86.2±4.5                       & 96.50±0.0                      & 96.60±0.0                      & 80.58±4.7                      & 85.60±2.9                       & 94.48±1.9                     & 95.05±2.0    \\                   
\textbf{NESS2}& 97.48±0.3                      & 97.44±0.2                       & 99.32±0.1                      & 99.27±0.1                        & 98.15±0.3                       & 97.85±0.1                       & 91.24±3.1                      & 90.72±4.1                        & \textbf{96.52±0.2}             & \textbf{96.67±0.1}              & 90.06±1.0                       & 90.17±1.9                       & \textbf{97.06±1.5}             & \textbf{97.24±1.4}  \\            
\textbf{NESS4}& 97.85±0.2                      & 97.76±0.2                       & \textbf{99.50±0.1}              & \textbf{99.43±0.1}              & \textbf{98.57±0.2}              & \textbf{98.13±0.2}              & 92.37±3.7                     & 92.52±3.3                       & 96.43±0.2                      & 96.60±0.1                       & 91.08±2.4                      & 92.06±3.4                       & 97.46±2.2                      & 97.13±2.5                      \\
\textbf{NESS8}& \textbf{97.93±0.2}             & \textbf{97.78±0.1}              & 99.21±0.1                      & 99.11±0.1                       & 98.26±0.3                      & 97.83±0.4                       & \textbf{94.16±1.7}             & \textbf{93.00±2.1}               & 96.05±0.2                      & 96.28±0.1                       & \textbf{95.46±3.1}             & \textbf{94.68±3.1}              & 97.55±0.6                      & 96.55±1.1     \\
\bottomrule

\end{tabular}}{}
}
\end{sc}
\end{small}
\end{center}
\vskip -0.1in
\end{table*}

\begin{figure*}[h]
\vskip 0.2in
\begin{center}
     \begin{subfigure}[c]{0.22\textwidth}
         \includegraphics[width=\textwidth]{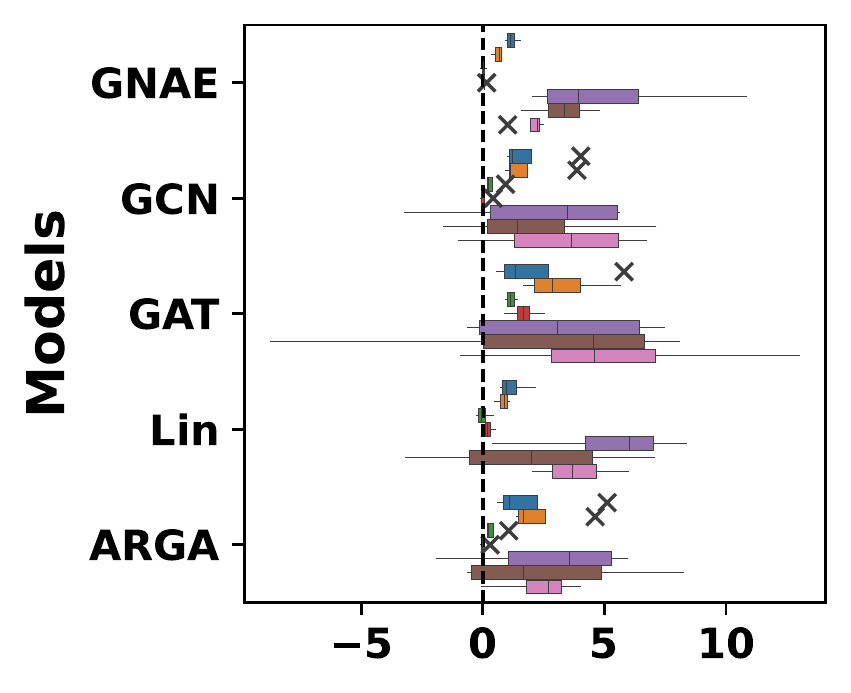}
         \caption{Aggregation}
     \end{subfigure}
     \begin{subfigure}[c]{0.22\textwidth}
         \includegraphics[width=\textwidth]{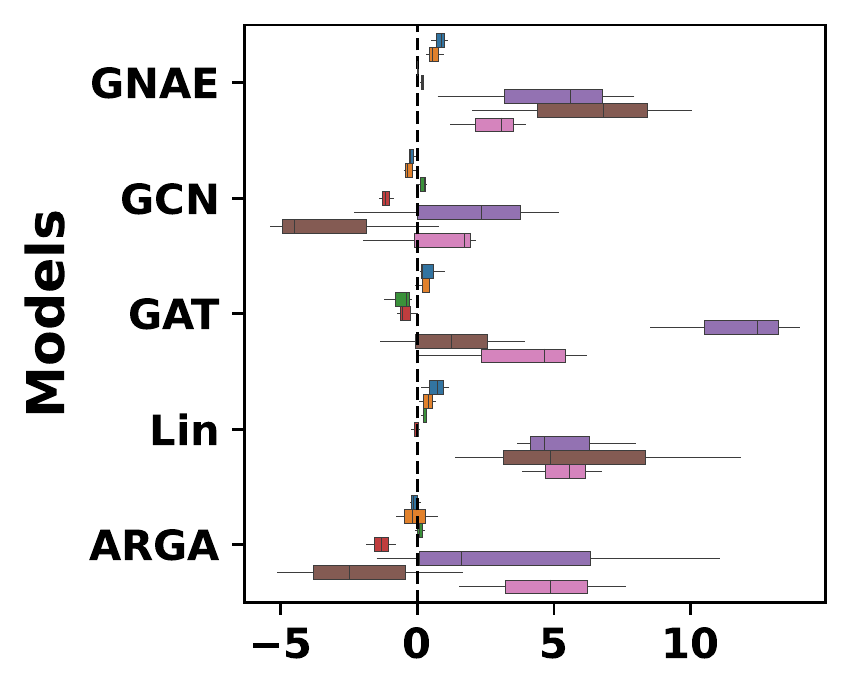}
         \caption{Reconstruction}
     \end{subfigure}
     \begin{subfigure}[c]{0.22\textwidth}
         \includegraphics[width=\textwidth]{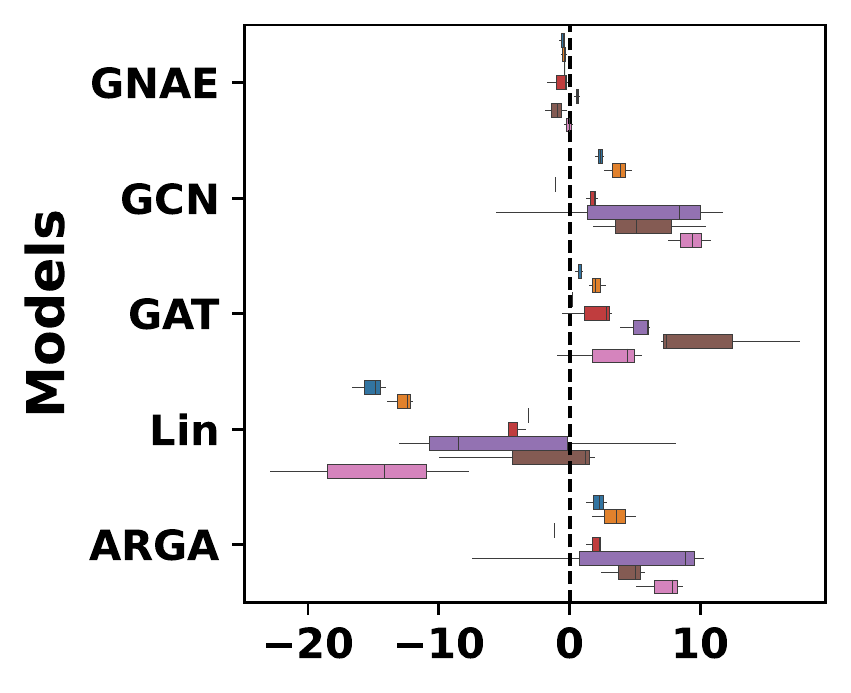}
         \caption{Contrastive loss}
     \end{subfigure}
     \begin{subfigure}[c]{0.31\textwidth}
         \includegraphics[width=\textwidth]{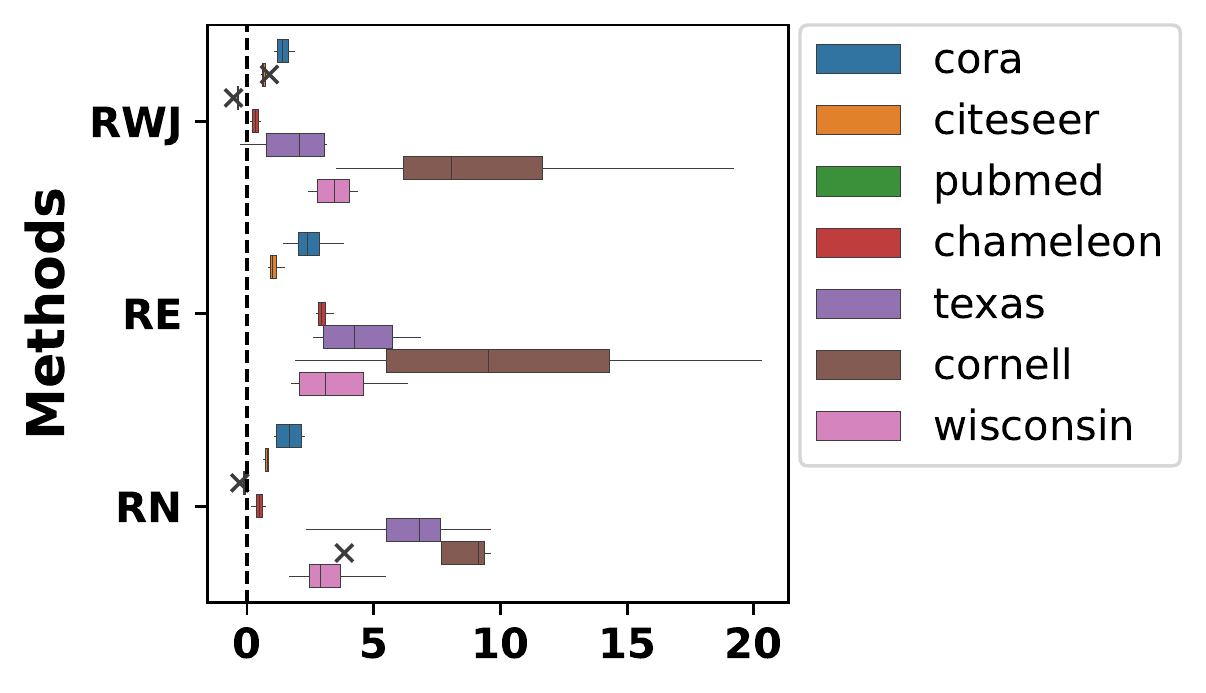}
         \caption{Sampling methods}
     \end{subfigure}
\caption{Experiments with NESS: \textbf{a)} The difference in AUC scores (\%) obtained in two ways for NESS2: i) Aggregating the latent variables of static subgraphs i.e. $\bm Z=Agg(\bm z_1, \bm z_2)$, (our default case), ii) Getting the latent variable of the entire graph directly i.e. $\bm Z=E(\bm X, \bm A)$ at test time. The aggregation performs better across all models and datasets; The difference in AUC scores (\%) when: \textbf{b)} we reconstruct subgraphs rather than full graph. \textbf{c)} we use contrastive and reconstruction loss rather than only reconstruction loss. \textbf{d)} comparing NESS2 to other sampling strategies.}\label{fig:ness_ablation1}
\end{center}
\vskip -0.2in
\end{figure*}

\textbf{The effect of aggregating latent variables in NESS}
We obtain the joint node embeddings of the graph by aggregating latent variables from static subgraphs. This approach is distinct from frameworks that utilize stochastic subgraphs, as their subgraphs are subject to variation in each iteration. Thus, in order to assess the efficacy of aggregating latent variables at test time, we measure the difference in performance between the use of aggregated node embeddings and direct encoding of the entire training graph (i.e. $AUC_{agg}-AUC_{direct}$). The results, displayed in Figure~\ref{fig:ness_ablation1}a, demonstrate a notable improvement in performance from the aggregation of subgraphs across various encoder types and benchmark datasets. We theorize that this gain comes from the ensemble of similarity scores used in NESS as described in Section~\ref{ness_principles}.

\textbf{Reconstructing subgraph versus large graph in NESS}
We compare performance of two training approaches: reconstructing subgraphs vs entire training graph. The results, displayed in Figure~\ref{fig:ness_ablation1}b, demonstrate the difference in performance ($AUC_{subgraph}-AUC_{full-graph}$) between the two approaches. We observe that reconstructing subgraphs often yields performance that is on par or superior to the alternative approach. 

\textbf{Applying contrastive learning in NESS}
We investigate the potential of contrastive learning to enhance performance in our setting. Figure~\ref{fig:ness_ablation1}c shows the difference in performance between using both reconstruction and contrastive losses rather than using only reconstruction (i.e. $AUC_{r+c}-AUC_{r}$). The results indicate that contrastive learning leads to a significant improvement in performance for GCN, GAT and ARGA models across all datasets while GNAE and Lin models experience a slight to severe degradation in performance. In light of results in Figure~\ref{fig:analysis_subgraphs}b, we posit that the subgraphs in GNAE and Lin settings are more in consensus, making them less effective in contrastive learning.

\textbf{Comparing NESS Sampling Strategies}
We use random edge splitting to partition the training graph into equally sized subgraphs without shared edges. We compare this with three other sampling approaches: i) \textit{Random edge (RE):} Edges are uniformly sampled randomly \citep{krishnamurthy2005reducing}; ii) \textit{Random walk with jump (RWJ):} Random walk sampling with probability p=0.1 to jump to any node \citep{leskovec2006sampling}; iii) \textit{Random node sampler (RN):} Nodes are uniformly sampled \citep{stumpf2005subnets}. In all settings, we sample two subgraphs, which might share edges and not cover the entire training graph. The joint embedding is still formed by aggregating latent variables at test time. The difference in performance between NESS2 and each setting (e.g., $AUC_{NESS}-AUC_{RE}$) is shown in Figure~\ref{fig:ness_ablation1}d for GNAE as the encoder. Our partitioning method provides a significant performance boost across datasets, but relative performance depends on the encoder and dataset as reported in Figure~\ref{fig:appx_comparing_sampling_approaches} in the Appendix.

\begin{figure}[t]
\vskip 0.2in
\begin{center}
     \begin{subfigure}[c]{0.33\columnwidth}
        \includegraphics[width=\columnwidth]{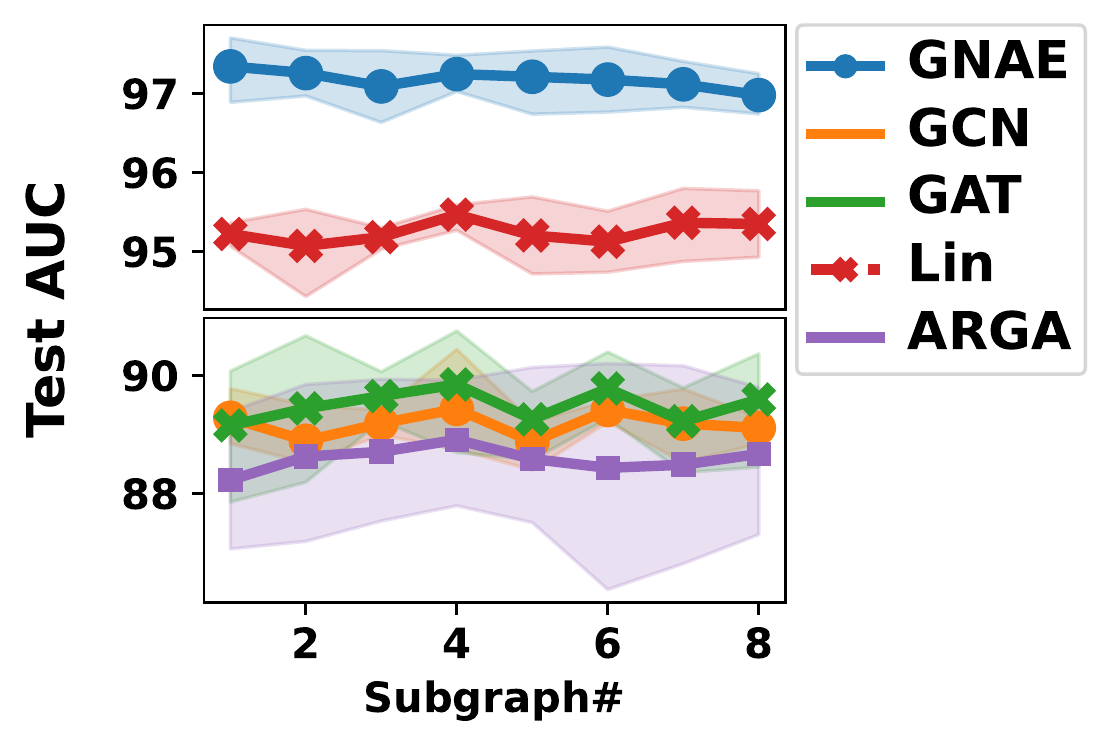}
         \caption{NESS8 - Cora}
     \end{subfigure}
     \begin{subfigure}[c]{0.31\columnwidth}
        \includegraphics[width=\columnwidth]{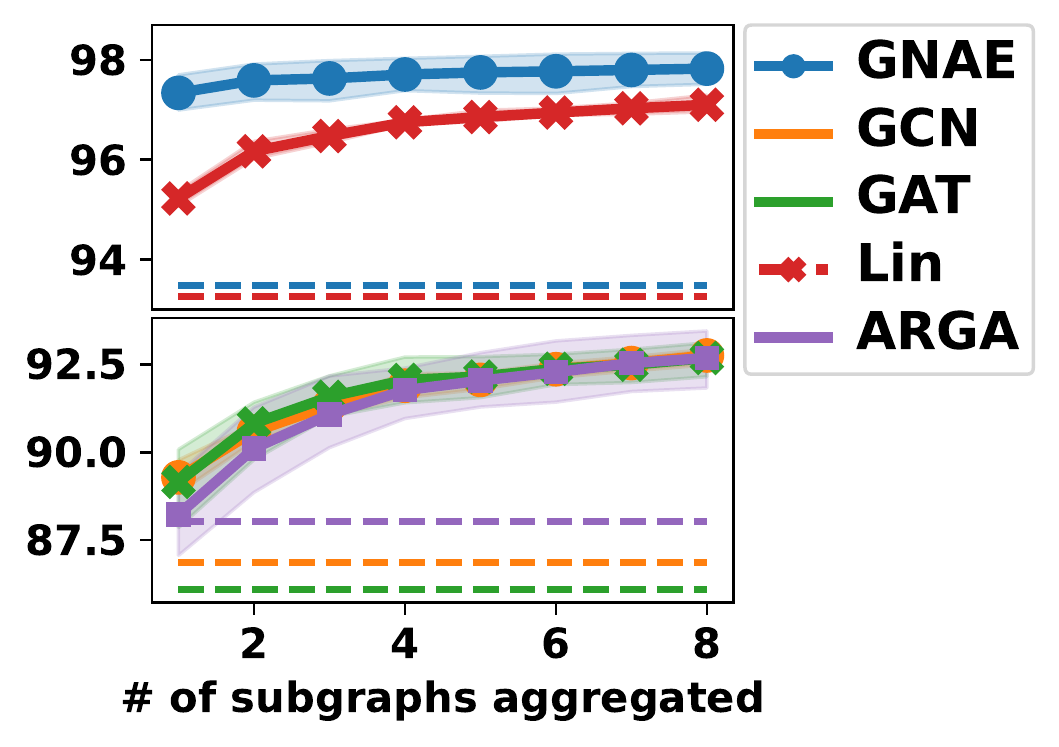}
         \caption{NESS8 - Cora}
     \end{subfigure}
     \begin{subfigure}[c]{0.32\columnwidth}
        \includegraphics[width=\columnwidth]{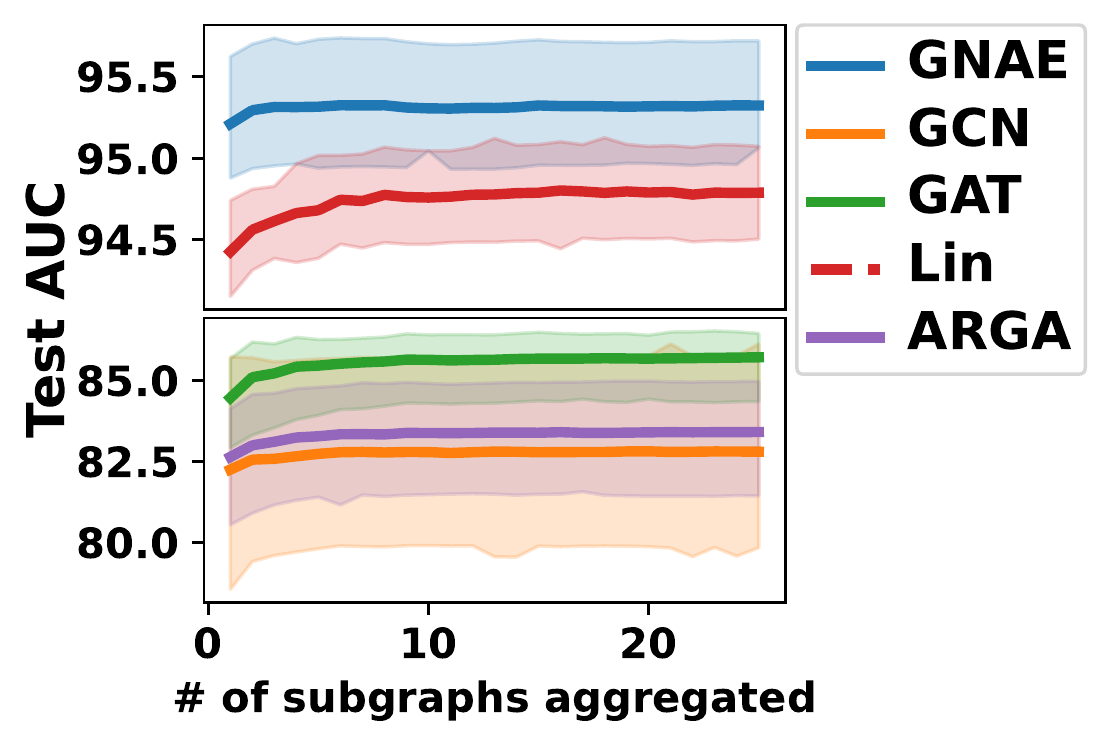}
         \caption{DRES2 - Cora}
     \end{subfigure}
\caption{\textbf{Evaluating the subgraphs of NESS8 (a, b) and of DRES2 (c) for Cora dataset:} \textbf{a)} The test AUC using the node embeddings of each subgraph. \textbf{b)} The test AUC using the mean aggregation of the latent representations of subgraphs, starting with the first subgraph (i.e. $\bm Z=\bm z_1$ at $x=1$), and keep adding new subgraphs to get a new joint node embedding sequentially (e.g., $\pmb{Z}=agg(\pmb{z_1}, ..., \pmb{z_4})$ at $x=4$). Dotted lines show the performance obtained by averaging the predictions of all subgraphs in all three cases ($K=2, 4, 8$) for each model, establishing the baselines in the form of the traditional ensemble models. \textbf{c)} Repeating the same experiment in (b) for DRES2.}\label{fig:gradual_aggregation_performance}
\end{center}
\vskip -0.2in
\end{figure}

\begin{wraptable}{R}{0.4\textwidth}
\centering
\vspace{-4mm}
\caption{Ablation study using Cora and Texas datasets with GNAE and GCN encoders; \textbf{RL/CL:} Reconstruction / Constrastive Loss, \textbf{Agg:} Aggregated embeddings, $\bm{\epsilon}$: Input noise (e.g., drop-edge).
\label{Tab:ablation}}
\resizebox{0.4\textwidth}{!}{
{\begin{tabular}{@{}c|c|c|c|c|c|c|c@{}}
\toprule {\bf RL} & {\bf Agg}  & {$\bm{\epsilon}$} & {\bf CL} &\multicolumn{4}{c}{\textbf{Test AUC}} \\\hline \hline
                  &         &                   &            & \multicolumn{2}{c|}{\textbf{Cora}} & \multicolumn{2}{c}{\textbf{Texas}} \\\cline{5-8}
                  &         &                   &            & \textbf{GNAE} & \textbf{GCN} &  \textbf{GNAE} & \textbf{GCN}  \\
\hline
{ +} & { - } & { - }  & { - }   & 93.22    & 76.95     & 90.78 & 68.28 \\
{ +} & { + } & { - }  & { - }   & 97.17 & 87.98 & 93.18  &  58.88 \\
{ +} & { + } & { + }  & { - }   & 98.13 & 92.25 & 94.68  & 65.29  \\
{ +} & { + } & { + }  & { + }   & \textbf{98.22} & \textbf{93.65} & \textbf{95.82}  & \textbf{82.54} \\
\hline
\end{tabular}}{}
}
\label{table:ablation}
\end{wraptable}

\textbf{Evaluating the predictive quality of node embeddings from each subgraph} In the NESS8 setting, we evaluate the effectiveness of the node embeddings from each of the eight subgraphs for link prediction in Figure~\ref{fig:gradual_aggregation_performance}a, discovering that some subgraphs are more informative than others, but the differences are not substantial. Additionally, we assess the quality of the joint embedding in a gradual manner in Figure~\ref{fig:gradual_aggregation_performance}b. As we progressively aggregate more latent variables of subgraphs to compute the joint embeddings, $\bm Z$, of the graph, we expect that $\bm Z$ becomes increasingly informative. Thus, starting with the first latent variable, {$\pmb{z}_1$}, as the entire graph's embedding, we compute the AUC score for link prediction on Cora's test set. We then sequentially add and aggregate each subsequent latent variable, up to $\pmb{z}_8$, to compute $\pmb{Z}$ for evaluation. As more subgraphs are aggregated, $\pmb{Z}$ becomes more expressive for the link prediction task, as illustrated in Figure~\ref{fig:gradual_aggregation_performance}b, leading to a significant performance boost.
We also conduct the same experiment for DRES2, which is trained using randomly sampled subgraphs during training. As shown in Figure~\ref{fig:gradual_aggregation_performance}c, aggregating the embeddings of randomly sampled subgraphs does not improve the performance much, indicating that node embeddings learned in randomly sampled subgraphs might be suffering from the issue of mixing of embeddings and high correlation, which is not desired as discussed in Section~\ref{ness_principles}. Overall, the homophilous graphs seem to benefit from both using sparse subgraphs (Figure~\ref{fig:gradual_aggregation_performance}a) and aggregation of latent variables (Figure~\ref{fig:gradual_aggregation_performance}b). Finally, we repeat these experiments for Texas dataset, which has a strong edge heterophily ratio, and report it in Section~\ref{aggregating_embeddings_texas} of the Appendix. We show that there are similar trends in heterophilous graphs although the impact of aggregating the latent vectors is less pronounced. This is not surprising, given that the ego-features are found to be the most important in heterophily setting while higher-order neighborhoods contribute the most in homophily \citep{zhu2020beyond}.

\textbf{Aggregating latents vs traditional ensemble approach} In Figure~\ref{fig:gradual_aggregation_performance}b, the dotted lines show the performance obtained by averaging the predictions of all subgraphs for each model. They establish the baselines in the form of the traditional ensemble approach. Our method of aggregating latent vectors before computing the link score gives significant performance boost compared to ensemble approach. This gain is due to having inter- and intra-subgraph dot products in NESS as discussed in Section~\ref{ness_principles}.

\textbf{Ablation study} We conduct an ablation study with a homophilous (Cora) and heterophilous (Texas) graph using GNAE and GCN encoders as shown in Table~\ref{Tab:ablation}. We use NESS4 for Cora and NESS8 for Texas. The results show that using aggregation, adding input noise, and using contrastive loss have additive contributions towards a better performance. We note that aggregating embeddings degrades performance for "Texas + GCN encoder" case. Adding contrastive loss recovers the performance loss. This phenomenon might be related to the role of the ego-embeddings in heterophilous graphs as discussed before.

\section{Related Work}


\textbf{Subgraphs and stochastic methods} Using subgraphs has been proposed to learn more expressive representations or to address scalibility issues with large graphs. For example, Sub2Vec \citep{adhikari2018sub2vec} proposes an unsupervised algorithm to learn feature representations of arbitrary subgraphs. There are also new proposals to improve the expressivity of  graph neural networks by introducing subgraph counting \citep{ bouritsas2021improving} and graph reconstruction theory \citep{cotta2021reconstruction}. And there are a number of proposals to address the scalibility issues associated with GNN-based encoders; \textbf{Cluster-GCN} \citep{chiang2019cluster} introduces a scalable GCN algorithm by designing batches based on a graph clustering algorithm such as METIS \citep{karypis1998fast} and extends GCN \citep{kipf2016semi} to inductive setting. The graph partitioning is done over the vertices in the graph in a way that the links within the clusters are much more than the ones between clusters since the error is proportional to the number of between-cluster links. During the training, the subgraph is sampled \textit{dynamically} from subset of clusters in each iteration. Cluster-GCN is an approximation to original GCN-algorithm, and can be used as an encoder in our framework. \textbf{GraphSaint} \citep{zeng2019graphsaint}, similarly to Cluster-GCN, starts each iteration with an independently sampled subgraph induced from  the nodes sampled by a pre-defined sampler. It then generates embeddings and compute loss for nodes in the subgraph. The authors use samplers such as random node sampler (RN), random edge sampler (RE), and random walk sampler (RWS). \textbf{GraphSage} \citep{hamilton2017inductive} extends GCNs into inductive setting while also proposing trainable aggregation functions. For a given node, it samples multiple neighborhoods with varying number of hops and search depth, and aggregates information from these neighborhoods using a set of aggregator functions. \textbf{FastGCN} \citep{chen2018fastgcn} addresses the scalibility issue of GCN by introducing importance sampling to sample the neighborhood of each node when aggregating their vector representations in forward pass. The most relevant work to ours is \textbf{FastGAE} \citep{salha2021fastgae} that proposes to use stochastic subgraphs for training GAEs in transductive learning setting. However, FastGAE encodes \textit{entire graph} as in standard GAE while using stochastic subgraphs when decoding. Subgraphs are sampled via node sampling with graph mining methods. The solution obtained at the end of FastGAE's training is only an approximation to that of standard GAE setting. Its main advantage is scalibility during training since it does not need to reconstruct the entire graph. 

\textbf{Contrastive learning} The most works applying constrastive learning assumes an inductive learning setting.  Common approaches to get the augmented view of a graph include adding or dropping nodes and edges randomly \citep{xu2021infogcl, you2020graph, zhu2020deep, papp2021dropgnn, rong2020dropedge}, shuffling nodes \citep{velickovic2019deep} and graph diffusion \cite{hassani2020contrastive}.

\section{Conclusion}
In this work, we introduce NESS, a novel GAE-based framework that utilizes \textit{static} subgraphs during training and aggregates their latent vectors \textit{at test time} to learn node embeddings in a transductive setting. We give insights about the design choices behind NESS by drawing some similarities and distinctions with ensemble learning. We additionally introduce a contrastive learning approach in transductive learning setting. NESS is shown to yield significant performance gains in link prediction tasks for small to medium-sized graphs. We leave further research on scalability, other graph-based tasks and extension to inductive setting as future work as they can introduce potential limitations to NESS. Lastly, while we use an autoencoder setting in this work, NESS can easily be adapted to a variational autoencoder setting.


\clearpage
\bibliography{main.bib}
\bibliographystyle{plainnat}

\newpage
\appendix

\renewcommand\thefigure{A\arabic{figure}}
\renewcommand\thetable{A\arabic{table}}
\setcounter{figure}{0} 
\setcounter{table}{0}

\section{Algorithm}

\begin{algorithm}[h]
   \caption{NESS}
   \label{alg:ness}
\begin{algorithmic}
   \STATE {\bfseries Input:} Graph $G=(V, E)$, adj. $\bm A$, node features $\bm X$
   \STATE {\bfseries Output:} Node representation $\bm Z$
   \STATE {\bfseries Initialize:} Encoder $Enc$
   \STATE Partition $G$ into $k$ static $subgraphs=[\bm a_1,\bm a_2, ...,\bm a_k]$ by using random edge split.
   \FOR{$epoch=1$ {\bfseries to} $max\_epocs$}
    \STATE $loss=0$, $z\_list=[]$
        \FOR{$a_k$ {\bfseries in} subgraphs}
            \STATE $\bm z_{k} = Enc(\bm X, \bm a_k)$  
            \STATE $\Hat{\bm a}_{k} = \sigma(\bm z_{k}\bm z_{k}^T)$  
            \STATE $loss = loss + recon\_loss(\bm a_{k},\Hat{\bm a}_{k})$  
            \STATE $z\_list.append(\bm z_{k})$  
        \ENDFOR
        \IF{contrastive loss == True}
            \STATE Get combinations of $\bm z_{k}$'s: $[(\bm z_{1}, \bm z_{2}), ...]$
            \STATE Compute contrastive loss $\mathcal{L}_c$, based on Eq.~\ref{eq4}
            \STATE $loss = loss + \mathcal{L}_c$
        \ENDIF
    \STATE Update encoder $Enc$
   \ENDFOR
    \STATE \bfseries  Return $\bm Z=aggregate(\bm z_{1},\bm  z_{2}, ...,\bm  z_{k})$

\end{algorithmic}
\end{algorithm}

\section{Details of Models}\label{details_of_models}

\textbf{GNAE:} We use a linear layers with hidden dimensions of 32 followed by $\mathcal{L}_2$ normalisation. We then generate final output by using approximate personalized propagation of neural predictions (APPNP) \citep{gasteiger2018predict} with $K=1$ and $\alpha=0$.

\textbf{GCN:} We use two convolutional layers with hidden dimensions of 64 and 32 respectively. We apply ReLU activation \citep{nair2010rectified} to the output of first layer.

\textbf{GAT:} The first layer consists of K = 8 attention heads with 8 features each, for a total of 64 features. We then apply an exponential linear unit (ELU) activation \citep{clevert2015fast}. The second layer has 32 units with K=1 attention head for all datasets, except Pubmed, for which we use K=8 as suggested in \citep{velivckovic2017graph}.

\textbf{Linear (Lin):} We use a single convolutional layer with 32 dimensions.

\textbf{ARGA:} We use the same architecture as GCN. We also use a discriminator network that has three linear layers with (64, 64, 1) dimensions. We apply ReLU activation to the first two layers of Discriminator.

\subsection{Implementation and resources}\label{implementation_resources}
We implemented our work using PyTorch \citep{NEURIPS2019_9015}. AdamW optimizer \citep{loshchilov2017decoupled} with $betas=(0.9, 0.999)$ and $eps=1e-07$ is used for all of our experiments. We used a compute cluster consisting of Tesla K80 GPUs throughout this work.

\section{Related Works (Continued)}

\textbf{Matrix factorization} There are numerous methods to extract node embeddings of a graph in a low-dimensional vector that captures the local structure and features of the nodes. The methods based on matrix factorization performs dimensionality reduction using a matrix, which can be an adjacency matrix \citep{ahmed2013distributed}, transition probability matrix \citep{cao2015grarep}, or Katz index \citep{katz1953new}. 

\textbf{Random Walks} These methods encode node embeddings by leveraging the random walk probability to explore the graph structure.  There are many ways to design such random walks. For example, we can treat the graph as a collection of short random walks \citep{perozzi2014deepwalk}, or can adjust the random walk to be a hybrid of the breadth first search (BFS) and the depth first search (DFS) \citep{grover2016node2vec}. Combining the random walk data with other models such as skip-gram \citep{mikolov2013efficient}, we can learn useful node representations.

\section{Additional Results}

\begin{figure*}[ht]
\vskip 0.2in
\begin{center}
     \begin{subfigure}[c]{0.21\textwidth}
         \includegraphics[width=\textwidth]{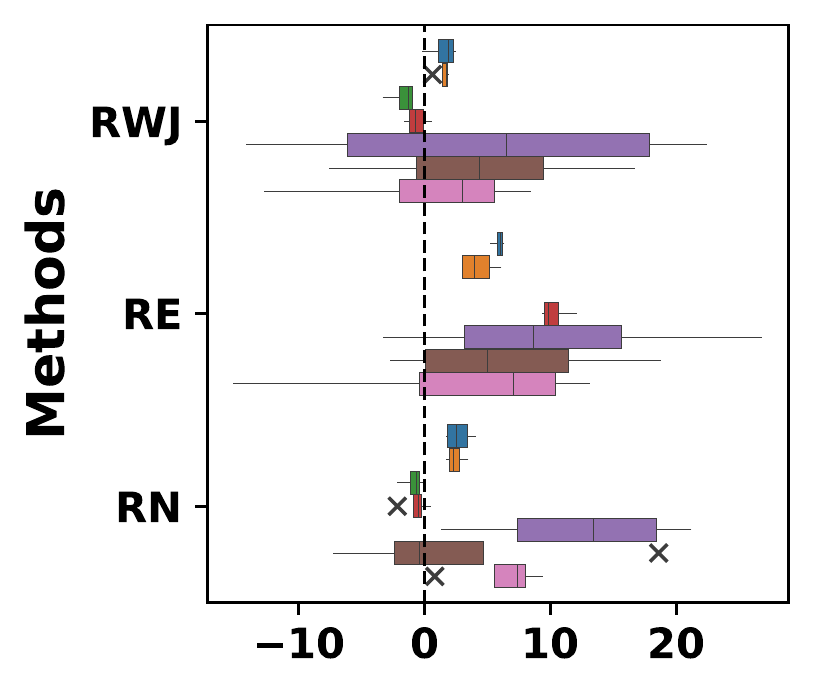}
         \caption{GAT}
     \end{subfigure}
     \begin{subfigure}[c]{0.21\textwidth}
         \includegraphics[width=\textwidth]{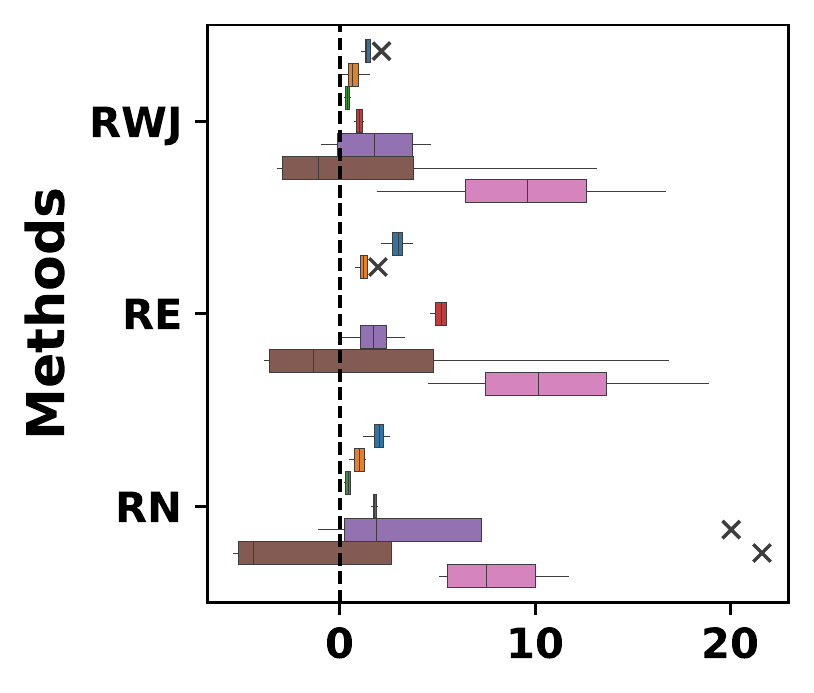}
         \caption{Lin}
     \end{subfigure}
     \begin{subfigure}[c]{0.21\textwidth}
         \includegraphics[width=\textwidth]{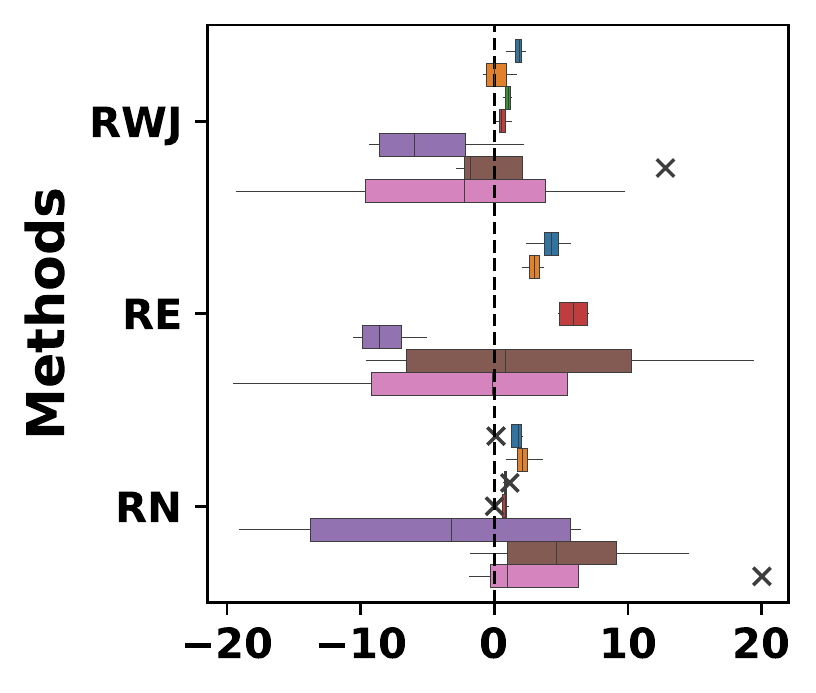}
         \caption{GCN}
     \end{subfigure}
     \begin{subfigure}[c]{0.3\textwidth}
         \includegraphics[width=\textwidth]{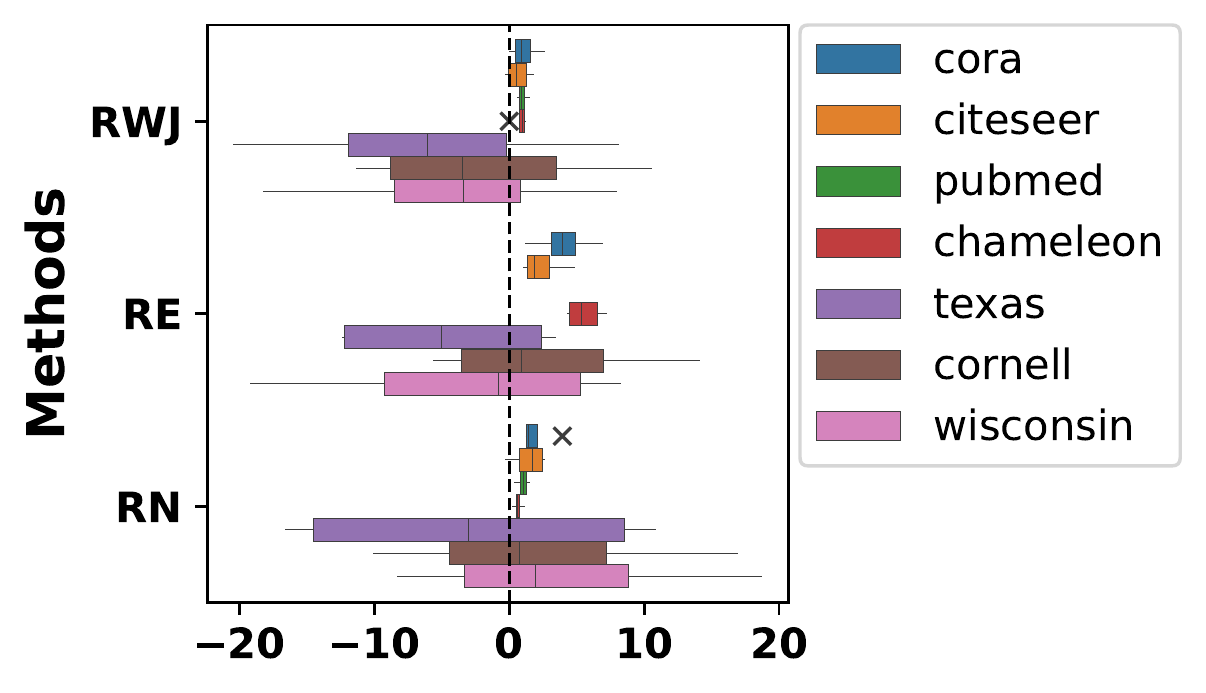}
         \caption{ARGA}
     \end{subfigure}
     
\caption{Showing difference in AUC scores (\%) that we obtain using our default method, compared to other sampling approaches applied during data preprocessing. Our default NESS setting using random edge split to partition the graph usually outperforms others.}\label{fig:appx_comparing_sampling_approaches}
\end{center}
\vskip -0.2in
\end{figure*}

Figure~\ref{fig:appx_comparing_sampling_approaches} shows the results for other models for the experiment reported in Figure~\ref{fig:ness_ablation1}d, comparing NESS to other settings with different sampling approaches either during data pre-processing or during training.

\subsection{Experiments with Texas dataset}\label{aggregating_embeddings_texas}

\begin{figure}[ht]
\vskip 0.2in
\begin{center}
     \begin{subfigure}[c]{0.32\columnwidth}
        \includegraphics[width=\columnwidth]{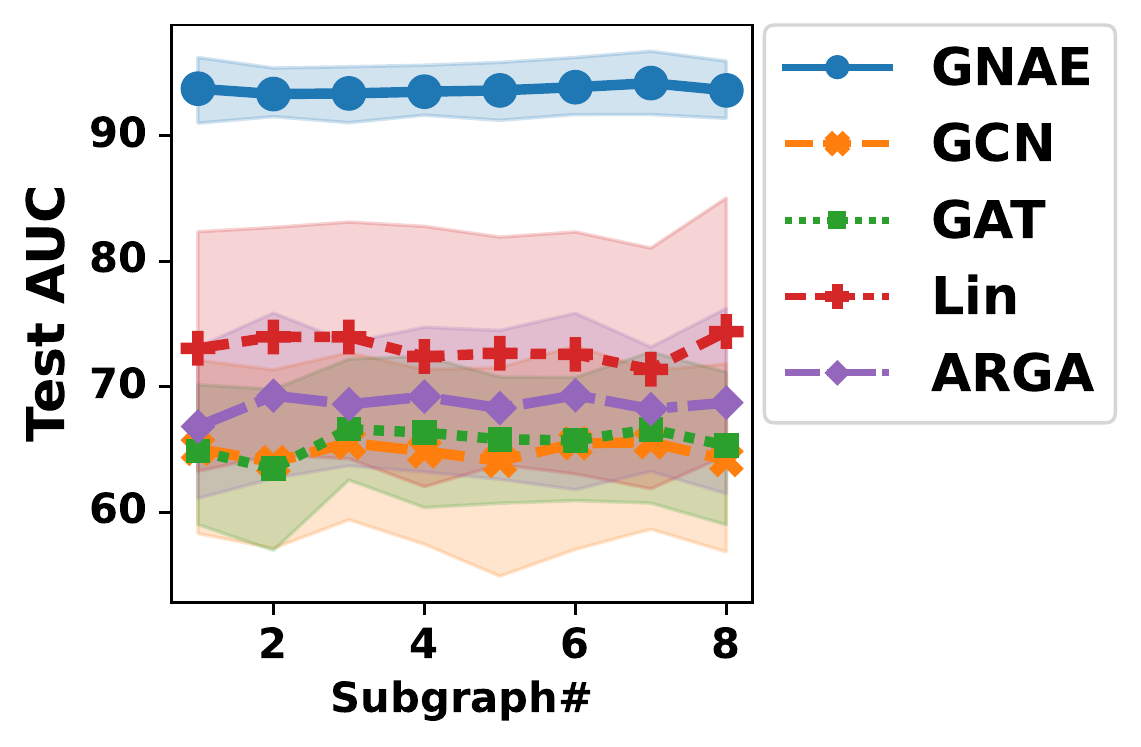}
         \caption{NESS8 - Texas}
     \end{subfigure}
     \begin{subfigure}[c]{0.29\columnwidth}
        \includegraphics[width=\columnwidth]{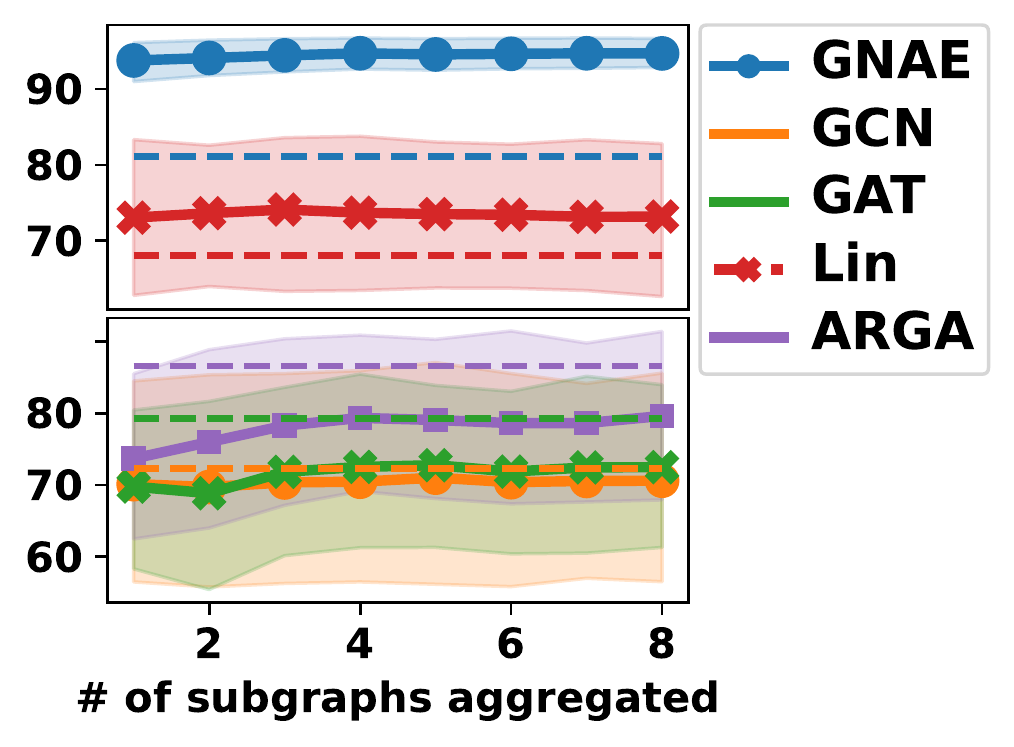}
         \caption{NESS8 - Texas}
     \end{subfigure}
     \begin{subfigure}[c]{0.32\columnwidth}
        \includegraphics[width=\columnwidth]{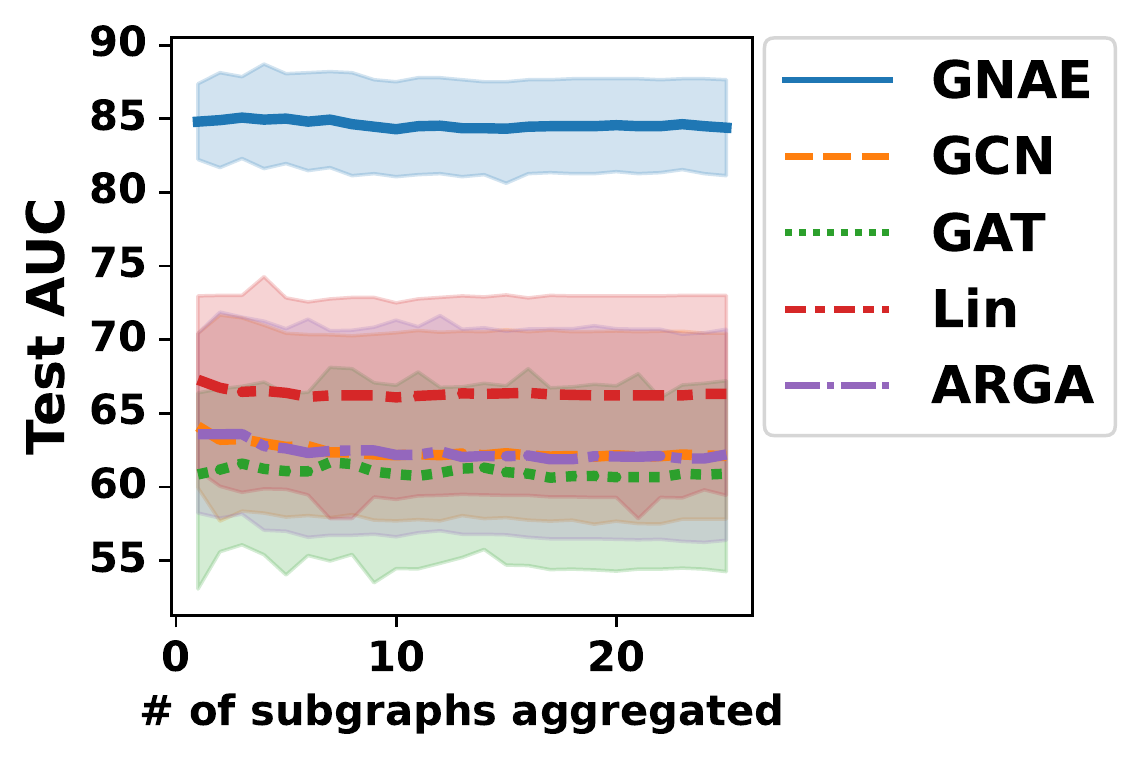}
         \caption{DRES2 - Texas}
     \end{subfigure}

\caption{\textbf{Evaluating the subgraphs of NESS8 (a, b) and of DRES2 (c) for Texas dataset:} \textbf{a)} The test AUC using the node embeddings of each subgraph. \textbf{b)} The test AUC using the mean aggregation of the latent representations of subgraphs, starting with the first subgraph (i.e. $\bm Z=\bm z_1$ at $x=1$), and keep adding new subgraphs to get a new joint node embedding sequentially (e.g., $\pmb{Z}=agg(\pmb{z_1}, ..., \pmb{z_4})$ at $x=4$). \textbf{c)} Repeating the same experiment in (b) for DRES2.}\label{fig:gradual_aggregation_performance_texas}
\end{center}
\vskip -0.2in
\end{figure}

\begin{figure}[h]
\vskip 0.2in
\begin{center}
     \begin{subfigure}[c]{0.32\columnwidth}
        \includegraphics[width=\columnwidth]{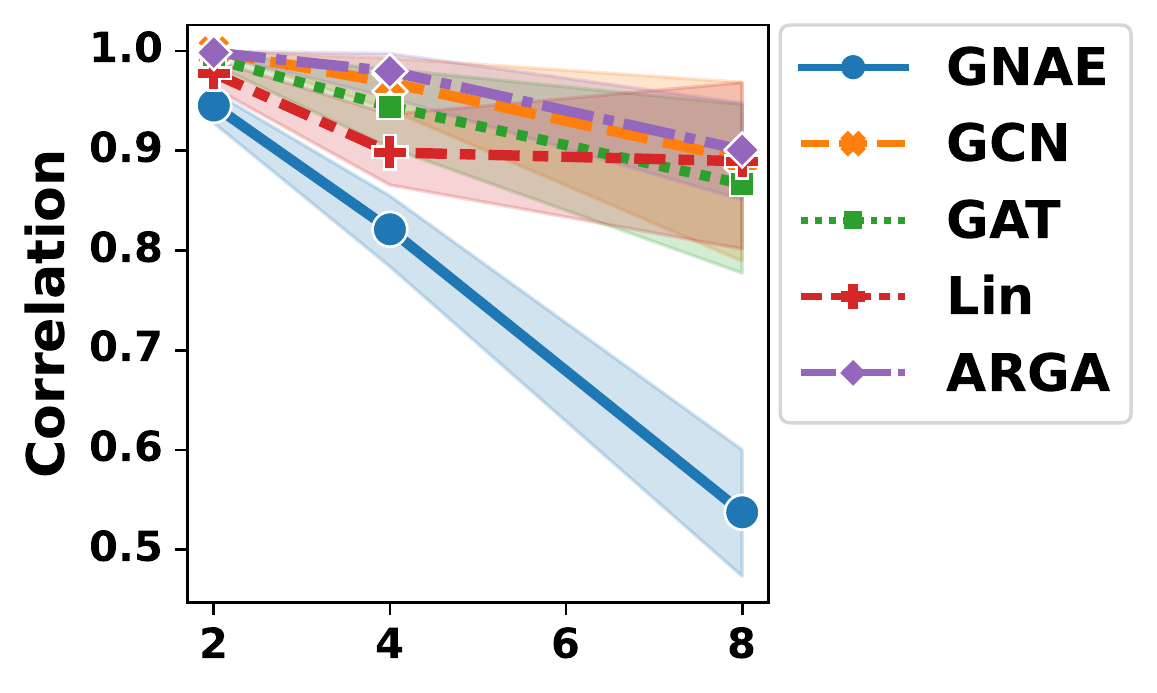}
         \caption{Correlation between subgraphs}
     \end{subfigure}
     \begin{subfigure}[c]{0.32\columnwidth}
        \includegraphics[width=\columnwidth]{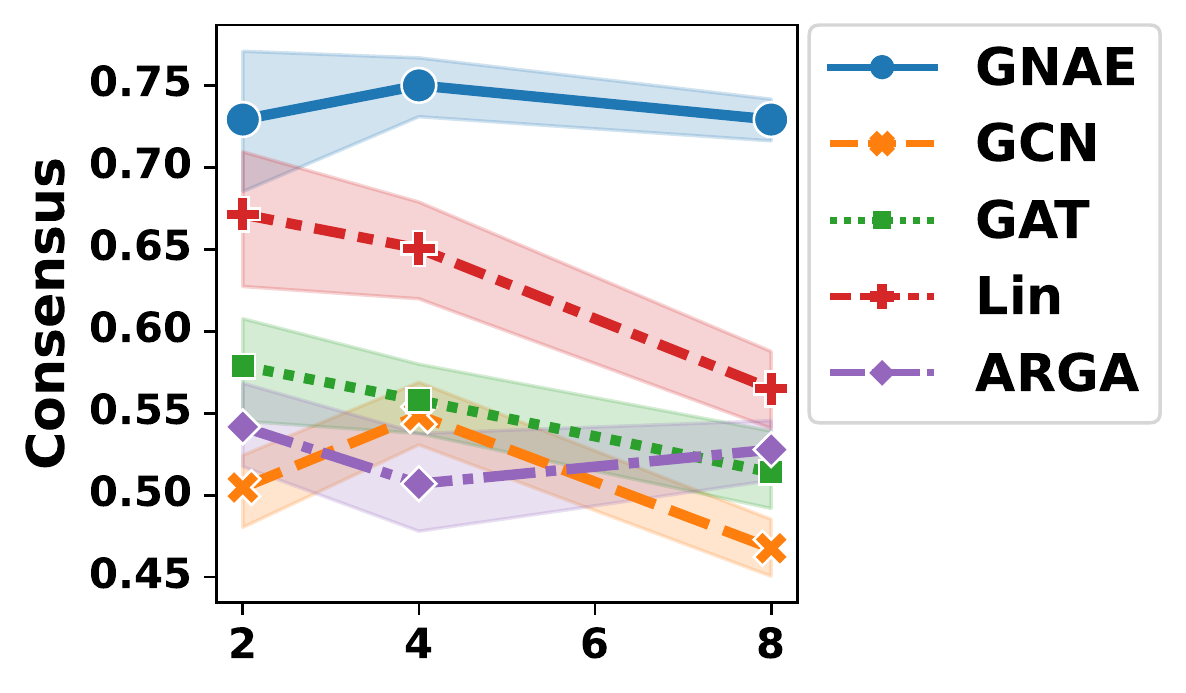}
         \caption{Consensus among subgraphs}
     \end{subfigure}
     \begin{subfigure}[c]{0.32\columnwidth}
        \includegraphics[width=\columnwidth]{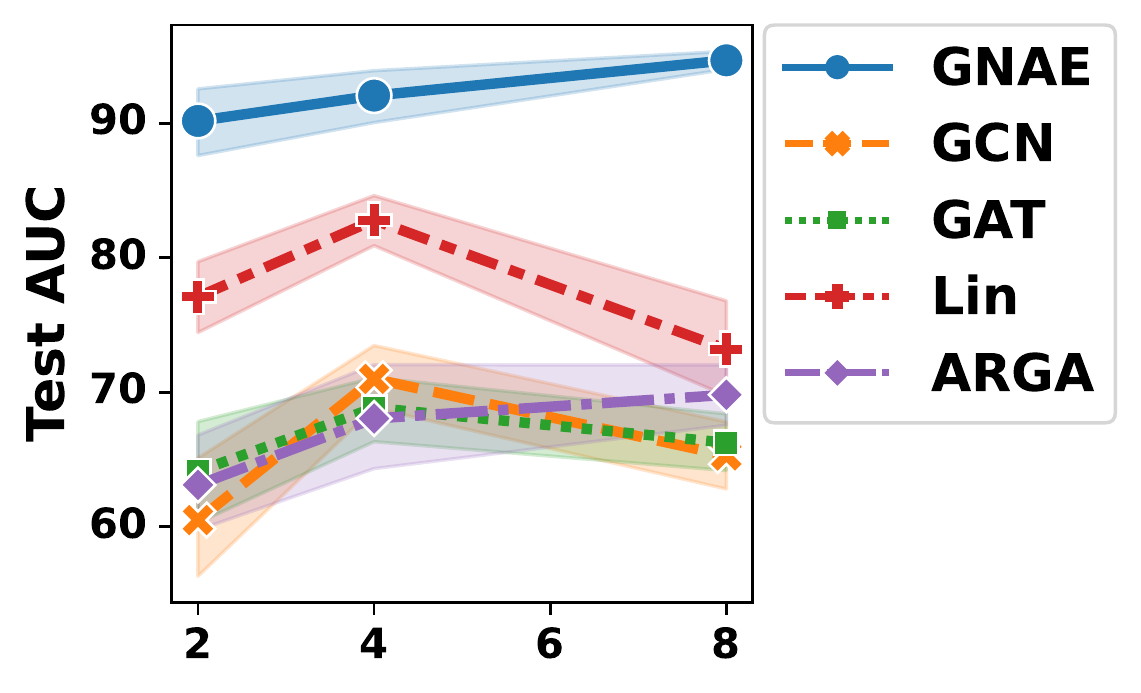}
         \caption{Test AUC using joint embedding}
     \end{subfigure}
\caption{\textbf{Analysis of subgraphs for Texas dataset:} \textbf{a)} Correlation between the representations of subgraphs,  \textbf{b)} Consensus: The ratio of links predicted correctly by all subgraphs in the test set,  \textbf{c)} Test AUC using the joint embeddings. X-axis corresponds to three cases; $K=2, 4$ and $8$.}\label{fig:analysis_subgraphs_texas}
\end{center}
\vskip -0.2in
\end{figure}

\section{Details of the Benchmark Datasets}

We list the details of three standard benchmark datasets from citation networks (Cora, Citeseer and Pubmed) \citep{sen2008collective}, three datasets (Cornell, Texas, Wisconsin) from WebKB \citep{pei2020geom},and Chameleon dataset \citep{rozemberczki2021multi} in Table~\ref{data_stats}.

\begin{table*}[h]
\caption{Dataset statistics. Homophily Ratios are taken from \citep{ma2021homophily}.}
\label{data_stats}
\vskip 0.15in
\begin{center}
\begin{small}
\begin{sc}
\resizebox{1.0\textwidth}{!}{
{\begin{tabular}{llllllll}
\toprule
\textbf{}              & \textbf{Cora} & \textbf{Citeseer} & \textbf{Pubmed} & \textbf{Chameleon} & \textbf{Cornell} & \textbf{Texas} & \textbf{Wisconsin} \\
\midrule
\textbf{Nodes}         & 2708          & 3,327             & 19717           & 2277               & 183              & 183            & 251                \\
\textbf{Edges}         & 5429          & 4732              & 44338           & 36101              & 295              & 309            & 499                \\
\textbf{Classes}       & 7             & 6                 & 3               & 5                  & 5                & 5              & 5                  \\
\textbf{Node Features} & 1433          & 3703              & 500             & 2325               & 1703             & 1703           & 1703               \\
\textbf{Homophily Ratio}           & 0.81         & 0.74             & 0.80            & 0.23              & 0.3            & 0.11          & 0.21     \\
\bottomrule
\end{tabular}}{}
}
\end{sc}
\end{small}
\end{center}
\vskip -0.1in
\end{table*}

\section{Broader Impact}\label{broader_impact}

Graph data has become a commonly used format in social media, healthcare, finance, law and many other fields. One of the important problems in graph domain is the task of finding missing links in the network. However, the performance of GNN-based models vary depending on the statistics and nature of the graph. Our paper proposes a method to boost the performance of most GNN-based encoders. The progress in this line of research will accelerate the adaptation of graph. However, we should be aware of the shortcomings of such adaptation in terms of biases and privacy issues that it might introduce.

\end{document}